\documentclass[authoryear, preprint, 3p]{elsarticle}
\usepackage{graphicx}
\usepackage{amsfonts}
\usepackage{amsmath,mathrsfs}
\usepackage{soul}
\usepackage{mathtools}
\usepackage{color}
\usepackage{graphicx,float,subfig,epstopdf}
\usepackage[capitalise]{cleveref}
\definecolor{mypink1}{rgb}{0.858, 0.188, 0.478}

\journal{Medical Image Analysis}

\begin{document}
\begin{frontmatter}

\title{Disease Prediction using Graph Convolutional Networks:\\Application to Autism Spectrum Disorder and Alzheimer's Disease}

\author[AB]{Sarah Parisot \corref{cor1}\fnref{fn1}}
\author[IMP]{Sofia Ira Ktena \fnref{fn1}}
\author[SINC]{Enzo Ferrante}
\author[IMP]{Matthew Lee} 
\author[SS]{Ricardo Guerrero\corref{cor2}}
\author[IMP]{Ben Glocker}
\author[IMP]{Daniel Rueckert}

\fntext[fn1]{Authors contributed equally.}
\cortext[cor1]{Corresponding author: S. Parisot, Aimbrain Solutions ltd, One Canada Square, London, E14 5AB. E-mail address: sarah@aimbrain.com. This work was carried out when the first author was affiliated with the Biomedical Image Analysis Group, Imperial College London, UK.}
\cortext[cor2]{This work was carried out when the author was affiliated with the Biomedical Image Analysis Group, Imperial College London, UK.}
%
%\authorrunning{Parisot et al.}   % abbreviated author list (for running head)
%

\address[IMP]{Biomedical Image Analysis Group, Imperial College London, UK }
\address[AB]{Aimbrain Solutions Ltd, London, UK}
\address[SINC]{Research Institute for Signals, Systems and Computational Intelligence, sinc(i), FICH-UNL/CONICET, Santa Fe, Argentina}
\address[SS]{StoryStream Ltd., London, UK}

\begin{abstract}
Graphs are widely used as a natural framework that captures interactions between individual elements represented as nodes in a graph. In medical applications, specifically, nodes can represent individuals within a potentially large population (patients or healthy controls) accompanied by a set of features, while the graph edges incorporate associations between subjects in an intuitive manner. This representation allows to incorporate the wealth of imaging and non-imaging information as well as individual subject features simultaneously in disease classification tasks. Previous graph-based approaches for supervised or unsupervised learning in the context of disease prediction solely focus on pairwise similarities between subjects, disregarding individual characteristics and features, or rather rely on subject-specific imaging feature vectors and fail to model interactions between them. In this paper, we present a thorough evaluation of a generic framework that leverages both imaging and non-imaging information and can be used for brain analysis in large populations. This framework exploits Graph Convolutional Networks (GCNs) and involves representing populations as a sparse graph, where its nodes are associated with imaging-based feature vectors, while phenotypic information is integrated as edge weights. The extensive evaluation explores the effect of each individual component of this framework on disease prediction performance and further compares it to different baselines. The framework performance is tested on two large datasets with diverse underlying data, ABIDE and ADNI, for the prediction of Autism Spectrum Disorder and conversion to Alzheimer's disease, respectively. Our analysis shows that our novel framework can improve over state-of-the-art results on both databases, with 70.4\% classification accuracy for ABIDE and 80.0\% for ADNI.

\end{abstract}


\end{frontmatter}

\section{Introduction}

Large scale collaborative initiatives and consortiums, like the Alzheimer's Disease Neuroimaging Initiative (ADNI), the Enhancing NeuroImaging Genetics through Meta-Analysis (ENIGMA) consortium~\citep{thompson2014enigma} and the International Neuroimaging Data-sharing Initiative (INDI)\footnote{\url{http://fcon\_1000.projects.nitrc.org/}}, acquire and share hundreds of terabytes of imaging, genetic, phenotypic and behavioural data in an effort to facilitate the discovery of novel biomarkers and better understand disease mechanisms. This ever-increasing volume of imaging and non-imaging information that is collected and shared among researchers stresses the need for computational models that are capable of representing potentially large populations, while exploiting all imaging modalities and additional data sources available.

Graphs provide a powerful and intuitive way of modelling individuals (as nodes) and associations or similarities between them (as edges). In this setting, a node can represent a specific modality of a subject's acquired data or a series of acquisitions at a particular time point, while the edge weights can be used to capture the similarities between each pair of nodes. Edges can also be non-weighted in this context, or binarised to indicate the absence or presence of association between two subjects/scans, which, however, limits their expressiveness. Graph-based models have been widely used for supervised (e.g. classification~\citep{tong2017multi}) and unsupervised tasks (e.g. manifold learning~\citep{wolz_nonlinear_2012,brosch_manifold_2013} and clustering~\citep{parisot_probabilistic_2016}) in population analysis, due to their capacity to accommodate complex pairwise interactions and ability to integrate non-imaging information. However, previous approaches have focused on summarising all available feature information via pairwise similarities and thus eliminated individual subject features and characteristics. For example,~\cite{tong2017multi} proposed a non-linear population graph fusion approach that aims to combine complementary multi-modal information about the subjects, while~\cite{zhao2014compact} focused on the graph construction and proposed an improved local reconstruction strategy for graph-based label propagation in Alzheimer's disease classification. On the contrary, methods that solely rely on imaging feature vectors~\citep{cuingnet2011automatic,abraham2016deriving} 
by training, for example, linear classifiers, are widely applied for classification analyses in large populations, but fail to capture interactions and similarities between subjects or their individual scans. A progressive method that seeks to learn an affinity matrix
from the observed imaging features while validating it on
the training data with known phenotype labels is proposed in~\citet{wang2017multi}, aiming to augment imaging with phenotypic information. Such analyses become more challenging and limited in performance when diverse imaging protocols are adopted for data acquisition within the same population, since they are more difficult to generalise.

\subsection{Graph Convolutional Neural Networks}
The advent of Convolutional Neural Networks (CNNs) as powerful models that exploit both image features (e.g. intensities) and spatial context by means of neighbourhood information (e.g. regular pixel grid) to yield hierarchies of features, has led to their application in numerous different problems related to 2D and 3D images, like image segmentation~\citep{havaei_brain_2017} and classification~\citep{hou2016patch}, long before their recent re-emergence~\citep{sahiner1996classification}. %more references
There is a direct analogy between an image segmentation task, where each pixel is to be assigned a label e.g. tissue of interest or background, and a subject classification task within a population e.g. for disease prediction. In the latter case, a subject along with its corresponding feature vector is equivalent to an image pixel with its intensity values for the different channels, while a graph constructed based on pairwise population similarities is equivalent to the pixel grid, for which the notion of proximity is more straightforward, since both describe the neighbourhood structure for convolutions. Nevertheless, the traditional widespread formulation of CNNs for regular domains cannot be directly extended to irregular ones. The description of the local neighbourhood structure and node ordering is not straightforward for irregular graph structures and these need to be properly defined to allow convolution and pooling operations~\citep{niepert2016learning}.

The first method dealing with neural networks on graphs was presented in \citet{scarselli2009graph}. In this pioneering work, the authors devised a mapping function from graph space to an $m$-dimensional Euclidean space, and proposed a supervised learning method to learn the parameters of their graph neural network (GNN) model. However, no convolution was considered in this model. The first work to introduce convolutional neural networks on graphs was described in \citet{bruna2013}. Bruna et al. used concepts from the emerging field of signal processing on graphs~\citep{shuman2013emerging}, giving rise to the spectral graph convolutional networks (GCNs). Spectral GCN is a generalisation of CNNs to irregular domains, which uses computational harmonic analysis to process signals observed on irregular graph structures~\citep{hammond2011wavelets}. The concepts borrowed from this field allow the extension of CNNs to irregular graphs in a principled way, by treating convolutions in the graph spatial domain as multiplications in the graph spectral domain. This kind of convolutions have several applications in computer vision, computer graphics and social network problems, among others, and have successfully been adopted to perform classification of documents in large citation datasets~\citep{kipf2016semi,levie2017cayleynets}.
Alternative approaches, where convolutions are directly defined in the spatial domain, have also been proposed in the literature. In \citet{masci2015geodesic}, for example, the authors define a local geodesic system of polar coordinates to extract “patches”, which are processed through a cascade of filters and activation functions. The so called geodesic convolutions are then used to construct graph convolutional networks which operate directly on the manifold.
Another recent work by~\citet{simonovsky2017dynamic} proposed to use filter weights conditioned on the edge labels, instead, and dynamically generate those for each input sample. This type of spatial convolutions on graphs are especially suitable for mesh structures, since they are local and allow to capture anisotropic patterns. A more detailed overview of such techniques is presented in~\citet{bronstein2017geometric}. However, spectral convolutions are preferential in cases where the underlying graph structure is fixed, as spatial graph CNNs tend to require more engineering. In this work, we use the convolution approach described in \citet{defferrard2016convolutional}, since it has shown outstanding performance for node classification tasks where the graph structure models different types of interactions between individuals within a population~\citep{kipf2016semi}.

\subsection{Graph-based Models for Disease Prediction}

Recently, the use of graph-based models has gained popularity in medical imaging applications, especially at a subject level.~\citet{kawahara2017brainnetcnn} used a customised version of CNNs on structural connectivity matrices to predict neurodevelopmental outcomes in preterm infants, which operates on the graph spatial domain and, therefore, captures only 1-hop neighbours in the receptive field of each node. As an alternative, spectral methods have been used to learn a similarity metric between functional connectivity networks~\citep{Ktena2017}, which was applied for Autism Spectrum Disorder (ASD) and sex classification as well as manifold learning~\citep{Ktena2018}. Spectral methods have also been explored for the prediction of visual tasks from MEG signals on a small number of subjects~\citep{guo2017deep}, while a bootstrapping strategy was used by~\citet{Anirudh2017} for ASD prediction.
Finally, \cite{lombaert2015spectral} combine spectral theory with random forests to process brain surfaces using spectral representations of meshes. Their proposed Spectral Forests are applied to the brain parcellation problem. 

In \citet{Parisot2017}, we proposed the first application of GCNs for group-level medical applications, more specifically for brain analysis in populations and diagnosis. We modelled populations as sparse graphs, where each node represents a subject and is associated with a feature vector extracted from imaging data. The edge weights encode the pairwise similarities between subjects and their features, and are obtained from auxiliary phenotypic data. This enables us to combine imaging and non-imaging data in a single framework. This population graph is fed as input to a GCN, which is trained in a semi-supervised manner from a subset of labelled nodes (e.g. of known diagnosis), aiming to classify the remaining, unlabelled nodes. The goal of the proposed method is to leverage the complementary non-imaging information available to explain the similarities between subjects within a graph structure and exploit the power of graph convolutions. Our main hypothesis is that integrating clinical expertise, i.e. non-imaging information that is known to be linked to specific pathologies, to model similarities between subjects can improve learned representations of image features and classification performance.  \\

\noindent \textbf{Contribution:} This paper constitutes an extended version of our work in \citet{Parisot2017}. In this work, we provide a deeper analysis of the method and modelling choices, through a substantially extended experimental evaluation as well as an in-depth discussion. We explore the influence of each main component of the model and provide discussion on the implications of the obtained results. This extensive evaluation is carried out on the ABIDE and ADNI databases, two large and challenging datasets, with the aim to diagnose Autism Spectrum Disorder (ABIDE) and to predict conversion from Mild Cognitive Impairment (MCI) to Alzheimer's Disease (ADNI). Our evaluation on two different datasets aims to demonstrate the framework's versatility as it facilitates the incorporation of domain-specific knowledge in two different clinical settings, while at the same time showing consistent improvement with respect to baselines for both challenging problems. 

The main contributions of the proposed work are:
\begin{itemize}
\item An introduction of GCNs for population analysis in the medical imaging domain.
\item A novel formulation of subject classification as a graph labelling problem, integrating imaging and non imaging data.
\item A seamless integration of known non-imaging features, allowing to integrate clinical expertise to boost classification performance. 
\end{itemize}
In addition, our extended version proposes the following enhancements with respect to the work introduced in \cite{Parisot2017}:
\begin{itemize}
\item A complete sensitivity analysis for key parameters of the proposed GCN model
\item New feature selection strategies for the ABIDE database
\item Detailed investigation of different graph structures and their influence on classification results
\item Comparison of our method to new baselines: Random Forest and Multi-Layer Perceptron classifiers
\item An improved, state of the art performance on both databases 
\end{itemize}

Our experiments show that exploiting GCNs with an accurate graph structure leads to significant improvements in classification accuracy, yielding state of the art results \cite{arbabshirani2017,abraham2016deriving, heinsfeld2018} for both ABIDE (70.4\% accuracy) % for degree=4
and ADNI (80\% accuracy). The model implementation is publicly available at \url{https://github.com/parisots/population-gcn}.

\section{Methods}

\begin{figure}[t]
\centering
\includegraphics[width=1\linewidth]{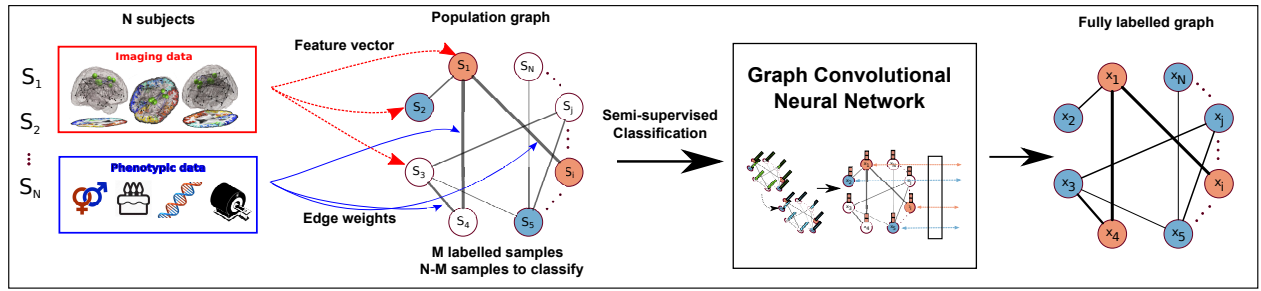}
\caption{Overview of the pipeline used for classification of population graphs using Graph Convolutional Networks.}
\label{fig:overview}
\end{figure}

An overview of the proposed method is shown in Fig. \ref{fig:overview}. We consider a population of $S$ subjects, each subject being described by/associated with a set of complimentary, phenotypic and demographic information (e.g. sex, age, acquisition site). The population comprises a set of $N$ imaging acquisitions (structural or functional MRI are considered in this paper), with $N \geq S$, meaning that one subject can be associated with several acquisitions (longitudinal scans). Our objective is to predict the status of each subject (healthy control or disease) from the imaging data informed by the phenotypic information. The population is represented as a weighted sparse graph, $\mathcal{G}=\{\mathcal{V},\mathcal{E}, W\}$, where $W$ is the adjacency matrix describing the graph's connectivity. Each acquisition $A_v$ is represented by a vertex $v \in \mathcal{V}$, corresponds to a subject $S_v$ and is associated with a $C$-dimensional feature vector $\mathbf{x}(v)$ extracted from the imaging data. The edges $\mathcal{E}$ of the graph model the similarity between the corresponding subjects and incorporate the phenotypic information.

We model our diagnosis task as a node classification problem, where we aim to assign a label $l \in \{0, 1\}$ to each graph node which describes the diseased ($l = 1$) or healthy status ($l = 0$) of the subject. Even though we focus on binary classification in this work, the model can easily be extended for multi-class classification problems. We adopt a semi-supervised strategy, where all node features along with the population graph are fed to the GCN, while only a subset of the graph nodes is labelled during training and used for the  optimisation process.  Intuitively, the graph acts as a regulariser, `encouraging' nodes connected with high edge weights to contribute more towards filtering the features of neighbouring nodes in a way that boosts label propagation performance.

\subsection{Databases and Preprocessing}
\label{subsec:dataset}

We demonstrate the potential and versatility of the model using two large and challenging databases, namely the \emph{ABIDE} and \emph{ADNI} databases. Each of the databases comprises subject specific information that can be leveraged through our population graph structure. 

The \textbf{ABIDE database} \citep{di2014autism} aggregates data from different international acquisition sites and openly shares neuroimaging (functional MRI) and phenotypic data of 1112 subjects\footnote{http://preprocessed-connectomes-project.org/abide/}. We select the same set of 871 subjects used by~\cite{abraham2016deriving} that met the imaging quality and phenotypic information criteria, comprising 403 individuals with ASD and 468 healthy controls. These subjects were regrouped based on location into 20 imaging sites. To ensure a fair comparison with the state of the art~\citep{abraham2016deriving}  we use the same preprocessing pipeline, the Configurable Pipeline for the Analysis of Connectomes (C-PAC) \citep{craddock2013}, which involves skull striping, slice timing correction, motion correction, global mean intensity normalisation, nuisance signal regression, band-pass filtering (0.01-0.1Hz). The functional images were registered to a standard anatomical space (MNI152) to allow cross-subject comparisons. 
Subsequently, the mean time series for a set of regions extracted from the Harvard Oxford (HO) atlas~\citep{desikan2006automated} were computed and normalised to zero mean and unit variance. The HO atlas distributed with FSL\footnote{http://www.fmrib.ox.ac.uk/fsl/} is based on anatomical landmarks and the version used in this work includes both cortical and subcortical (excluding left/right WM, left/right GM, left/right CSF and brainstem) ROIs, yielding 111 regions in total. 
The individual connectivity matrices $M_1, ..., M_N$ are estimated by computing the Fisher transformed Pearson's correlation coefficient between the representative rs-fMRI timeseries of each ROI in the HO atlas.  The correlation matrices are, then, Fisher transformed to improve normality.

The \textbf{ADNI database} is the result of efforts from several academic and private co-investigators \footnote{http://adni.loni.usc.edu}. To date, ADNI in its three studies (ADNI-1, -GO and -2) has recruited over 1700 adults, aged between 55 and 90 years, from over 50 sites from the U.S. and Canada. 
In this work, a subset of 540 early/late MCI subjects that contained longitudinal T1 MR images and their respective anatomical segmentations was used. Our inclusion criteria were therefore: MCI diagnosis, available longitudinal T1 MR images and available corresponding segmentations as described in \cite{Ledig2015}. MCI often represents an intermediate stage between normal cognition and Alzheimer's disease. Therefore, the conversion from MCI to AD is more challenging to predict than distinguishing between HC and AD patients. In total, 1675 samples were available, with 289 subjects (843 samples) diagnosed as AD at any time during follow-up and labelled as converters. It should be noted that the AD diagnosis in the dataset comes from clinical evaluation only and is missing histopathological confirmation. As a result, the AD diagnosis should be referred to as ``probable AD''. For simplification purposes, we use the term AD to refer to probable AD in the remainder of this paper. Longitudinal information ranged from 6 to 96 months, depending on each subject. Acquisitions after conversion to AD were not included.  
As of $1^{st}$ of July 2016 the ADNI repository contained 7128 longitudinal T1 MR images from 1723 subjects. ADNI-2 is an ongoing study and therefore data is still growing. Therefore, at the time of a large scale segmentation analysis (into 138 anatomical structures using MALP-EM \citep{Ledig2015}) only a subset of 1674 subjects (5074 images) was processed, from which the subset used here was selected. 

With the \emph{ABIDE database}, we aim to separate healthy controls from patients suffering from ASD.  This database comprises data acquired at different sites and using different protocols, which results in a highly heterogeneous database where the acquisition protocol can strongly affect the comparability of subjects. Our goal with the \emph{ADNI database} is to predict whether a patient with MCI will convert to AD, in other words, we aim to separate patients with stable MCI from those with progressive MCI. Our strategy is to intrinsically model/exploit the longitudinal aspect of the data within our graph structure, so as to highlight the importance and advantage of modelling the interactions between different input data.

\subsection{Population graph construction}
\label{subsec:graph_construction}

We provide an illustration of the construction of the population graph in Fig.~\ref{fig:overview}. This graph construction is a key aspect of the method, as an inappropriately constructed graph (i.e. a graph that does not accurately explain the similarity between subjects and their feature vectors) will fail to exploit the power of GCNs. Very inaccurate graph structures could even worsen performance compared to a simple linear classifier. Intuitively, this would equate to performing image convolutions on unrelated pixels (e.g. that are randomly spread across the image) instead of a local image patch. %I like this example 
The two main decisions required to build the population model are : 1) the definition of the feature vector $\mathbf{x}(v)$ describing each graph node/acquisition, and 2) the connectivity of the graph, i.e. its edges $\mathcal{E}$ and their weights $W$, which models the similarity between nodes/subjects/scans and their corresponding features.

\subsubsection{Feature vector}

We extract the feature vector purely from imaging data, as would typically be the case for image based classification. Our objective is to demonstrate how classification using simple image features can be enhanced using complementary information in the graph structure. 

For the ADNI dataset, we use the volumes of all $C=138$ segmented brain structures, a type of feature which has been highly effective for prediction of Alzheimer's disease, due to the impact of the disease on the brain's structure and volume differences between healthy, MCI and AD populations \citep{ries2008magnetic}.

There is increasing evidence that ASD is linked to disruptions in the functional and structural organisation of the brain ~\cite{abraham2016deriving,rudie2013altered}. As a result, we use functional connectivity derived from resting-state functional Magnetic Resonance Imaging (rs-fMRI) for ASD classification using the ABIDE dataset. More specifically, we use the vectorised functional connectivity matrices, i.e. the upper triangular elements of the square adjacency matrices, as feature vectors. This simple approach has had numerous successes for fMRI based classification, it was notably used in \cite{abraham2016deriving}, setting the state of the art on the whole ABIDE dataset at 67\% with a simple linear classifier. 
Due to the high dimensionality of the connectivity matrix, we explore, alongside using the whole feature vector, different dimensionality reduction strategies to input a $C$ dimensional feature vector to the network. Our different strategies, detailed in section \ref{sec:feat_selec}, are recursive feature elimination using a ridge classifier, a simple autoencoder, a multilayer perceptron classifier and principal component analysis. It is worth noting that feature selection is not used for ADNI data, which has a much smaller and tractable feature vector size.

% The proposed model requires two critical design choices: 1) the definition of the feature vector $\mathbf{x}(v)$ describing each sample, and 2) modelling the interactions between samples via the definition of the graph edges $\mathcal{E}$.
% We keep the feature vectors simple so as to focus on evaluating the impact of integrating contextual information in the classification performance. For the ABIDE data-set, we follow the method adopted by  \cite{abraham2016deriving} and define a subject's feature vector as its vectorised functional connectivity matrix. Due to the high dimensionality of the connectivity matrix, a ridge classifier is employed to select the most discriminative features from the training set. For the ADNI dataset, we simply use the volumes of all 138 segmented brain structures. 

\subsubsection{Graph edges}

Similarly to pixel neighbourhood systems, the graph structure provides a broader field of view, filtering the value of a feature with respect to its neighbours' instead of treating each feature individually. It has to be carefully crafted so as to accurately model the interactions between feature vectors. 
Our hypothesis is that non-imaging complementary data can provide key information to explain the associations between subjects' feature vectors. The objective is to leverage this information, in order to define an accurate neighbourhood system that optimises the performance of the subsequent graph convolutions.
Therefore, it is important to select the phenotypic measures which best explain similarities between the imaging data, or similarities between the subjects' labels. 

Considering a set of $H$ non-imaging phenotypic measures $\mathbf{M} = \{M_h\}$ (e.g. subject's sex, or age), the population graph's adjacency matrix $W$ is defined as follows: 
\begin{equation}
W(v,w)= Sim(A_v,A_w) \sum_{h=1}^H \gamma(M_h(v),M_h(w)),
\end{equation}
\noindent where, $Sim(S_v,S_w)$ is a measure of similarity between subjects, increasing the edge weights between the most similar graph nodes; $\gamma$ is a measure of distance between phenotypic measures.

$\gamma$ is defined differently depending on the type of phenotypic measure integrated in the graph. For categorical information such as subject's sex, we define $\gamma$ as the Kronecker delta function $\delta$, meaning that the edge weight between subjects is increased if e.g. they have the same sex. Constructing edge weights from quantitative measures (e.g. subject's age) is slightly less straightforward. In such cases, we define $\gamma$ as a unit-step function with respect to a threshold $\theta$: 
\begin{equation}
\gamma(M_h(v),M_h(w)) = \begin{cases} 1 &\mbox{if } \vert M_h(v)-M_h(w) \vert < \theta  \\ 
0 & \mbox{otherwise.} \end{cases}
\end{equation}  

Both ADNI and ABIDE databases provide an extensive list of phenotypic features. In this work, we select three different phenotypic measures for each database that are considered relevant to the corresponding disease. 
For ABIDE, we consider \emph{acquisition site}, \emph{sex} and \emph{age} as our three potential phenotypic measures. The ABIDE dataset is highly heterogeneous due to the fact that the data has been acquired at different sites, using diverse imaging protocols and scanners at each location. As a result, the imaging data is best comparable between feature vectors acquired at the same site, making acquisition site an essential non-imaging measure to introduce. Sex and age are also considered, since sex differences have been observed in several studies on ASD suggesting that females are affected less frequently than males~\citep{werling2013sex}, while there are age-related group differences in functional connectivity overall~\citep{kana2014brain}.
%ira{age is mentioned here but we didn't use it in the MICCAI version - should we keep it?}\sarah{a reviewer asked why we don't use age. Since I added a new measure for ADNI, I thought we could do the same for ABIDE, all experiments have been done so i don't see the advantage of removing it,:}
%\sarah{sex/age ASD predisposition?}
Finally, we define the similarity measure as 
\begin{equation}
Sim(A_v,A_w) = \exp\left(-\frac{[\rho(\mathbf{x}(v),\mathbf{x}(w))]^2}{2 \sigma ^ 2}\right),
\end{equation}  
where $\rho$ is the correlation distance and $\sigma$ determines the width of the kernel. The idea behind this similarity measure is that subjects belonging to the same class (healthy or ASD) tend to have more similar networks (larger $Sim$ values) than subjects from different classes.

The ADNI graph is also built using the subject's \emph{sex} and \emph{age} information. These measures are chosen because our feature vector comprises brain volumes, which can strongly be affected by age and sex.  We also integrate genetic information, namely the presence of the APOE $\epsilon$4 allele, known to be a major risk factor for the development of Alzheimer's disease~\citep{mosconi2004mci}.  

Last but not least, the $Sim(A_v,A_w)$ function plays a particularly important role in the ADNI set-up. It is designed here to leverage the longitudinal information, strongly highlighting that several acquisitions correspond to the same subject. While linear classifiers treat each entry independently, here we define:
\begin{equation}
Sim(A_v,A_w) =  \lambda \text{ with } \begin{cases} \lambda > 1 &\mbox{if } S_v = S_w  \\ 
\lambda =0 & \mbox{otherwise.} \end{cases}
\end{equation}
This measure indicates the strong similarity between acquisitions of the same subject. 

The influence of each phenotypic and similarity measure will be investigated in our experiments section, so as to optimise the structure of our phenotypic graph. It should be noted that while handedness constitutes an important phenotypic measure associated to autism subtypes~\cite{soper1986handedness}, we could not integrate it in this study as this information was missing for a large number of subjects.

\subsection{Feature Selection Strategies}
\label{sec:feat_selec}

As mentioned in Sec. \ref{subsec:graph_construction}, the feature vector chosen for ABIDE classification (the vectorised fMRI connectivity network) has a high dimensionality (particularly with respect to the graph size of 871 nodes) which can negatively impact the performance of the algorithm and lead to overfitting issues. 
In \cite{Parisot2017}, we used a ridge classifier to perform \textbf{Recursive Feature Elimination} (RFE) with a fixed number of features $C$. It is an iterative process, where each iteration trains the ridge classifier on the training set with the current feature vector. The classifier's coefficients are used to sort the importance of the features and prune the ones with the smallest coefficient (i.e. the least discriminative) from the feature vector. This process iterates until the desired number of features is obtained. In this paper, we investigate 3 additional approaches as well as the influence of $C$. 

First, we use the well known \textbf{Principal Component Analysis} (PCA), which is very commonly used for dimensionality reduction purposes, using singular value decomposition to project the data to a space of lower dimensionality. PCA is not very adapted to our problem due to the much larger dimension of the feature vector compared to the number of samples (6105 vs 871).

\begin{figure}[t]
\centering
\includegraphics[width=1\linewidth]{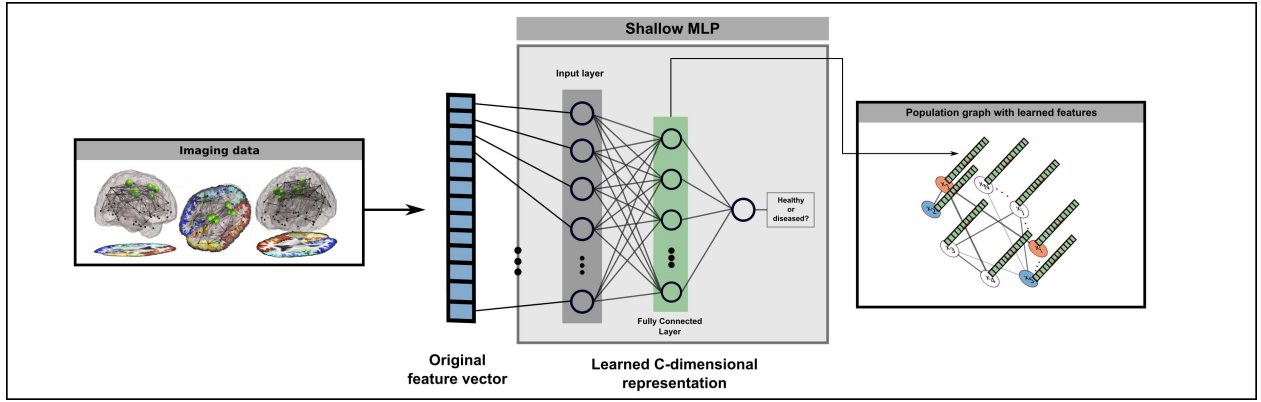}
\caption{Illustration of the MLP approach for ABIDE feature selection}
\label{fig:mlp}
\end{figure}

Second, we use two simple models based on artificial neural networks, one supervised and one unsupervised.
The first approach is a \textbf{Multilayer Perceptron} (MLP), a supervised feedforward neural network which consists of one hidden layer of size $C$. It is trained as a classifier using the ABIDE training data. The feature vector is obtained by extracting the $C$ dimensional feature vector obtained at the MLP's hidden layer. The underlying idea is that the MLP will learn a representation of the data specifically for the classification task, similar to what is done using the RFE approach. However, even if we are using a shallow MLP with a single hidden layer, there is a strong possibility of overfitting due to our limited amount of training data. The MLP model is illustrated in Fig.\ref{fig:mlp}.

\begin{figure}[t]
\centering
\includegraphics[width=1\linewidth]{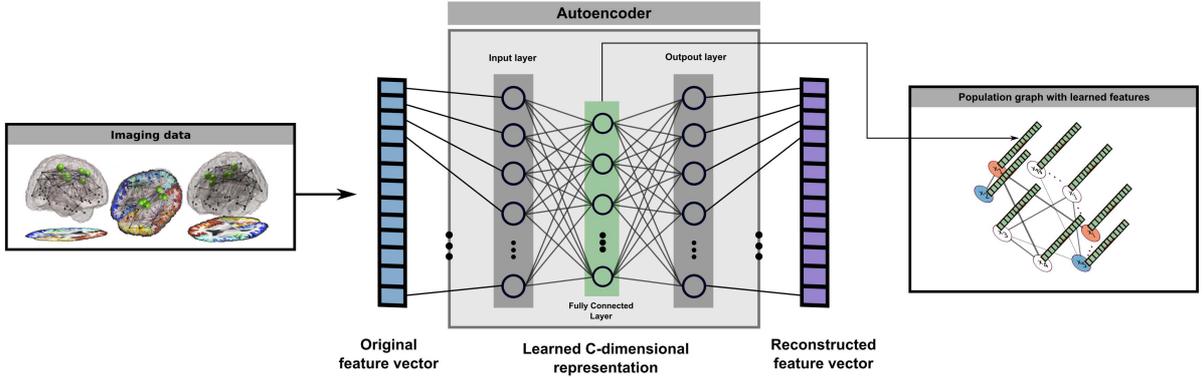}
\caption{Illustration of the Autoencoder approach for ABIDE feature selection}
\label{fig:autoencoder}
\end{figure}

\textbf{Autoencoders} are unsupervised neural networks that aim to learn a lower dimensional representation (a code) of an input data. It is made of an encoder (which learns the code) and a decoder (which reconstructs the input). It is trained by comparing the reconstructed input to the original data. Our autoencoder has tied weights and comprises one single hidden layer of size $C$ with a sigmoid activation and a tanh activation at the output layer. We use the mean square error as a loss function for training. The autoencoder model is illustrated in Fig.\ref{fig:autoencoder}.

\subsection{Graph Labelling using Graph Convolutional Neural Networks}
\label{subsec:filtering}
In the previous section we described the construction of the population graph based on phenotypic measures, and feature selection strategies on the original imaging data. In this section, we present the concept of spectral graph convolutions that serves as building block for the final graph convolutional neural network model, and discuss the GCN's architectural details.

\subsubsection{Spectral graph convolutions}

The discretised convolutions commonly used in computer vision intrinsically exploit the regular grid-like structure of e.g. 2D or 3D images. As a result, their generalisation/application to irregular graphs is not straightforward. We propose to use a recent formulation relying on spectral theory and graph signal processing~\citep{shuman2013emerging}.

% Discretised convolutions, those commonly used in computer vision, cannot be easily generalised to the graph setting, since these operators are only defined for regular grids, e.g. 2D or 3D images. Therefore, the definition of localised graph filters is critical for the generalisation of CNNs to irregular graphs. This can be achieved by formulating CNNs in terms of spectral graph theory, building on tools provided by graph signal processing (GSP)~\cite{shuman2013emerging}.

Spatial graph convolutions can be computed in the Fourier (spectral) domain as multiplications. The concept of graph Fourier transform (GFT) is introduced in \citet{shuman2013emerging} by analogy with the Euclidean domain, as an expansion of the Laplace operator in terms of its eigenfunctions. The normalised graph Laplacian of a weighted graph $\mathcal{G} = \{ \mathcal{V}, \mathcal{E}, W\}$ is defined as $\mathcal{L} = I_N - D^{-1/2} W D^{-1/2}$ where $I_N$ and $D$ are respectively the identity matrix of size $N \times N$ and the diagonal degree matrix. 
For any signal $\mathbf{x} \in \mathbb{R}^N$ the graph Laplacian acts as a difference operator and yields
\begin{equation}
(\mathcal{L}x)(i) = \sum_{j \in \mathcal{N}_i} W_{ij}(x(i) - x(j)),
\end{equation}
\noindent with $\mathcal{N}_i$ denoting the neighbours connected to vertex $i$ by an edge.

% The concept of spectral graph convolutions exploits the fact that convolutions are multiplications in the Fourier domain. The graph Fourier transform is defined by analogy to the Euclidean domain from the eigenfunctions of the Laplace operator. The normalised graph Laplacian of a weighted graph $\mathcal{G} = \{ \mathcal{V}, \mathcal{E}, W\}$ is defined as $\mathcal{L} = I_N - D^{-1/2} W D^{-1/2}$ where $I_N$ and $D$ are respectively the identity matrix of size $NxN$ and the diagonal degree matrix. 
% For any signal $\mathbf{x} \in \mathbb{R}^N$ the graph Laplacian acts as a difference operator and yields
% \begin{equation}
% (\mathcal{L}x)(i) = \sum_{j \in \mathcal{N}_i} W_{ij}[x(i) - x(j)],
% \end{equation}

 An eigendecomposition of the Laplacian matrix, $\mathcal{L}=U \Lambda U^T$, gives a set of orthonormal eigenvectors $U=[u_0,...u_{N-1}] \in \mathbb{R}^{N \times N}$ with associated real, non-negative eigenvalues $\Lambda=diag([\lambda_0, ..., \lambda_{N-1}]) \in \mathbb{R}^{N \times N}$. The eigenvectors associated with low frequencies/eigenvalues vary slowly across the graph, meaning that vertices connected by an edge of large weight have similar values in the corresponding locations of these eigenvectors.

% Its eigendecomposition, $\mathcal{L}=U \Lambda U^T$, gives a set of orthonormal eigenvectors $U \in \mathbb{R}^{N \times N}$ with associated real, non-negative eigenvalues $\Lambda \in \mathbb{R}^{N \times N}$. The eigenvectors associated with low frequencies/eigenvalues vary slowly across the graph, meaning that vertices connected by an edge of large weight have similar values in the corresponding locations of these eigenvectors. 

Considering a spatial signal $\mathbf{x}$ defined on graph $\mathcal{G}$, its Fourier transform is defined as $\hat{\mathbf{x}} \doteq U^T\mathbf{x} \in \mathbb{R}^{N}$, while the inverse transform is given by $\mathbf{x} \doteq U \hat{\mathbf{x}}$. 
A spectral convolution of signal $\mathbf{x}$ with a filter $g_{\theta}=diag(\theta)$ defined in the Fourier domain can then be defined as a multiplication in the Fourier domain: 
\begin{equation}
g_{\theta} \ast \mathbf{x} = g_{\theta}(\mathcal{L})\mathbf{x} = g_{\theta}(U \Lambda U^T)\mathbf{x} = U g_{\theta}(\Lambda) U^T \mathbf{x},
\end{equation}
\noindent where $\theta \in \mathbb{R}^{N}$ are the parameters of filter $g_{\theta}$.
% The graph Fourier Transform (GFT) of a spatial signal $\mathbf{x}$ is defined on the graph $\mathcal{G}$ as $\hat{\mathbf{x}} \doteq U^T\mathbf{x} \in \mathbb{R}^{N}$, while the inverse transform is given by $\mathbf{x} \doteq U \hat{\mathbf{x}}$. Using the above formulations, spectral convolutions of the signal $\mathbf{x}$ with a filter $g_{\theta}=diag(\theta)$ are defined as 
% $g_{\theta} \ast \mathbf{x} = g_{\theta}(\mathcal{L})\mathbf{x} = g_{\theta}(U \Lambda U^T)\mathbf{x} = U g_{\theta}(\Lambda) U^T \mathbf{x}$,

 Following the work of~\citet{defferrard2016convolutional}, we restrict the class of considered filters to polynomial filters $g_{\theta}(\Lambda) = \sum_{k=0}^{K}\theta_k \Lambda^k$. This approach has two main advantages: 1) it yields filters that are strictly localised in space (a $K$-order polynomial filter is strictly $K$-localised) and 2) it significantly reduces the computational complexity of the convolution operator. Indeed, such filters can be well approximated by a truncated expansion in terms of Chebyshev polynomials which can be computed recursively. 
 
It should be noted that additional simplifications on graph convolution layers are proposed in \citet{kipf2016semi} with applications to citation networks. Nonetheless, this simpler approach was not adapted to our relatively smaller and more complex datasets, which required the less efficient, yet more powerful and expressive approach, described in \citet{defferrard2016convolutional}. 

\subsubsection{GCN model}

\begin{figure}[t]
\centering
\includegraphics[width=1\linewidth]{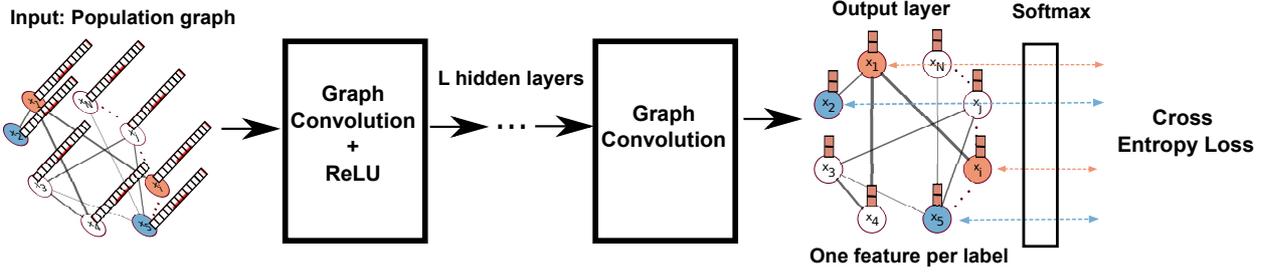}
\caption{Architecture of our GCN model.}
\label{fig:archi}
\end{figure}

Our model architecture is illustrated in Fig. \ref{fig:archi}. The model is relatively simple and consists of a fully convolutional GCN with $L$ hidden layers activated using the Rectified Linear Unit (ReLU) function. The output layer is followed by a softmax activation function. 
The graph is trained using the whole population graph as input. The training set comprises a labelled subset of graph nodes on which the loss function is evaluated and gradients are back propagated. The test set features (remaining unlabelled graph nodes) are observed during training, and influence the convolutions of labelled samples, making this a semi-supervised classification scheme. We further use a cross entropy loss function for the optimisation process. After training the GCN model, the softmax activations are computed on the test set, and the unlabelled nodes are assigned the labels maximising the softmax output.

\section{Results}

\subsection{Experimental set-up}

We evaluate our model on both the ADNI and ABIDE databases using a 10-fold stratified cross validation strategy. The use of 10 folds facilitates the comparison with the ABIDE state of the art \citep{abraham2016deriving} where a similar strategy is adopted. To provide a fair evaluation for ADNI, we ensure that all longitudinal acquisitions of the same subject are in the same fold (i.e. either the testing or training fold). 
GCN parameters that are not explored in this paper are fixed and chosen according to \cite{Parisot2017}, where they were optimised for the tasks considered here using a grid search. For ABIDE, we use: $L=1$, dropout rate: 0.3, l2 regularisation: $5.10^{-4}$, learning rate: 0.005, epochs: 150. The parameters for ADNI are: $L=6$, dropout rate: 0.02, l2 regularisation: $1.10^{-5}$, learning rate: 0.01, epochs: 200. Finally, graph construction variables $\lambda$ for ADNI similarity and $\theta$ for quantitative phenotypic measures are $\lambda = 10$ and $\theta=2$. 
In this section, we refer to the graph constructed from phenotypic data, as described in Sec. \ref{sec:graph_cons}, as the \emph{phenotypic} graph.  
The default structures for experiments using the \emph{phenotypic} graph are the ones used in \cite{Parisot2017}. The phenotypic measures used to build edges are \emph{SEX} and \emph{AGE} for ADNI, and \emph{SEX} and \emph{SITE} for ABIDE. Similarly, the default polynomial order is set to $K=3$.

\subsection{ABIDE Feature Selection Strategy}

\begin{figure}[t]
\centering
\includegraphics[width=0.45\linewidth]{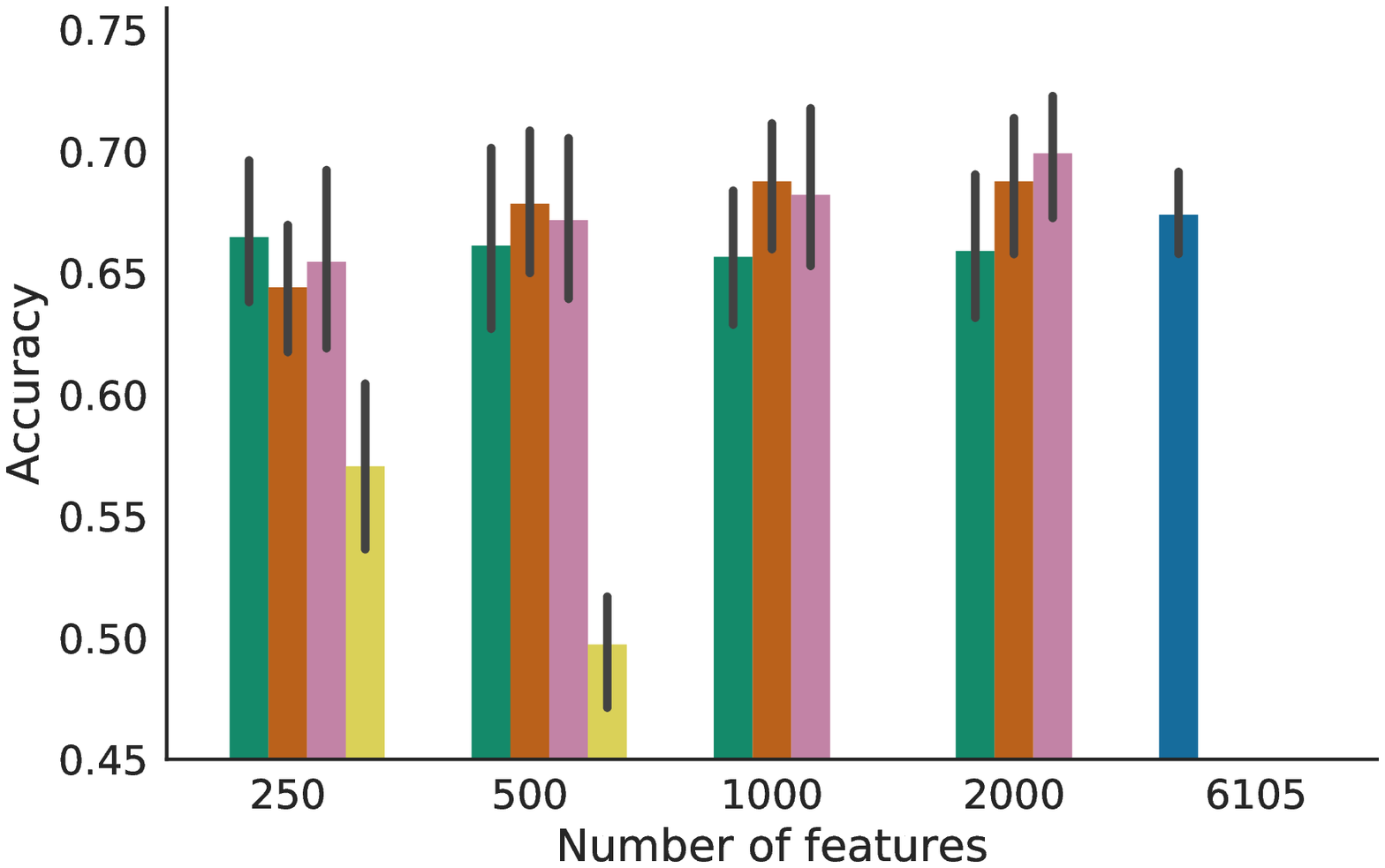}
\includegraphics[width=0.5\linewidth]{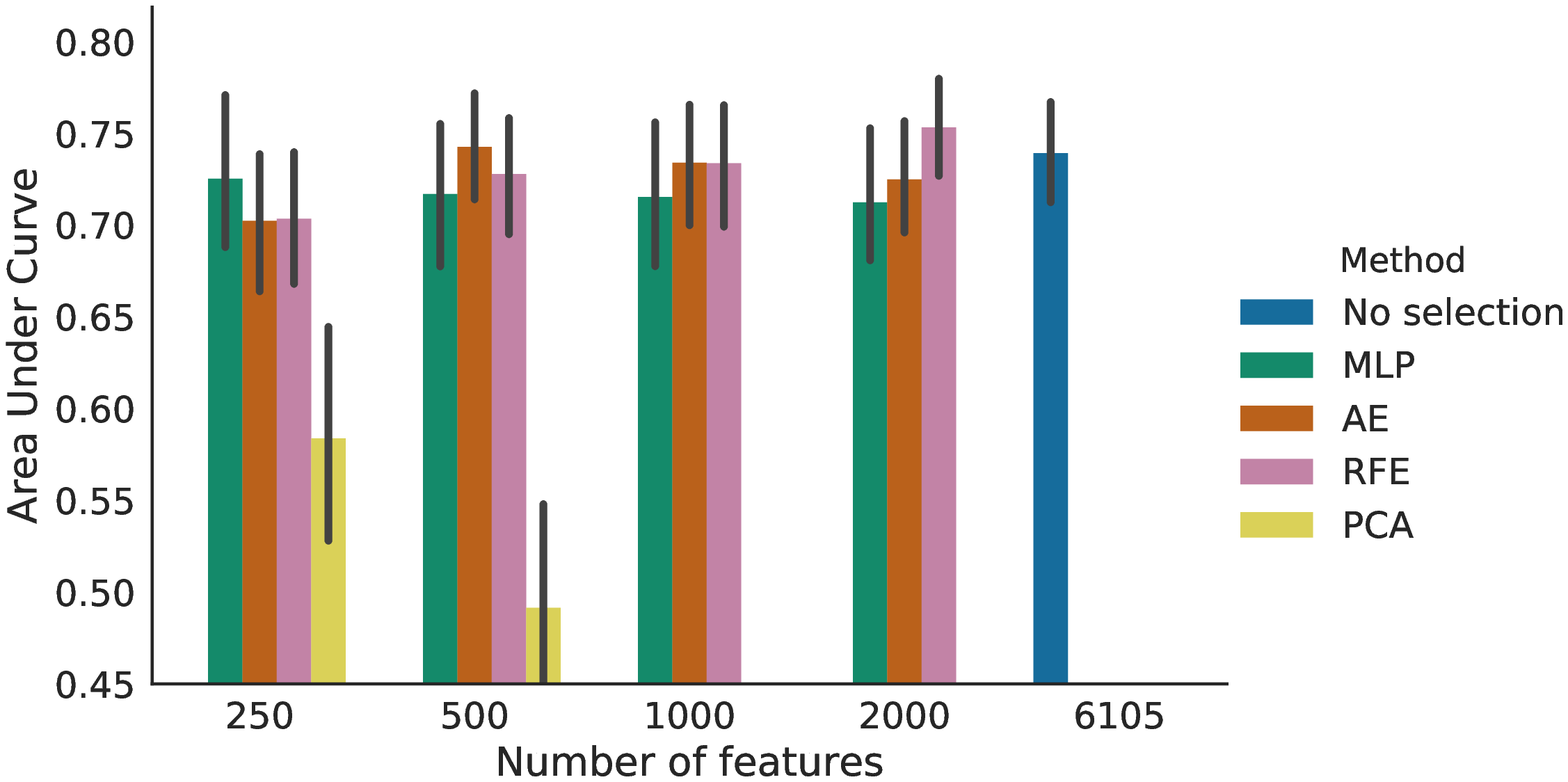}
\caption{ABIDE classification accuracy (a) and area under curve (b) for different feature selection strategies and number of features. The error bars report the classification accuracy across the 10 different folds. }
\label{fig:fselec}
\end{figure}

We report the influence of the feature selection scheme on the ABIDE database for the four considered methods described in Sec.~\ref{sec:feat_selec}, i.e. RFE, PCA, MLP and Autoencoder (AE). We train and evaluate a GCN using RFE, MLP and Autoencoder for $C=\{ 250, 500, 1000, 2000\}$ number of features. For PCA, we report results for $C=\{ 250, 500\}$ which correspond to approximatively $85\%$ and $95\%$ of explained variance. Last but not least, we report results for $C=6105$, which corresponds to using the whole feature vector as input. 
AE is trained for 100 epochs with a learning rate of $5e^{-4}$, while the scikit-learn \citep{sklearn} implementation with default parameters is used for MLP and PCA. 

Results are shown in Fig. \ref{fig:fselec}, reporting classification accuracy as well as Area Under Curve (AUC) for the 10 different cross validation folds. 
The first observation is the very poor performance of PCA as a feature selection strategy. As mentioned in Sec. \ref{sec:feat_selec}, this approach, although very popular and efficient for the dimensionality reduction task, is limited to finding a linear projection of the features to the lower-dimensional space. Therefore, it is not as expressive as a non-linear method, like the Autoencoder. Additionally, since the ABIDE dataset is highly heterogeneous, the variance captured in the training set is likely not as representative of the test set.

MLP has the best performance with respect to other methods for 250 features but drops in performance when more features are added. This can be explained by the high tendency of the MLP classifier to overfit in our particular set up (notably due to our limited number of input samples). Features are learned from the training set, therefore optimised for classification of this specific subset of the data, which in turn, leads to overfitting when training the GCN model. 

Autoencoders do not have this tendency, and our AE model has the best performance for 500 and 1000 features. RFE obtains the overall best performance for $C=2000$ features, and its performance increases as the number of features increase. The fact that AE is performing better at lower resolution can be explained by the fact that RFE loses more information (every time the number of features is reduced, the RFE eliminates new elements of the feature vector) while AE provides a more compact representation of the whole feature vector, reducing therefore information loss with respect to RFE. 
The better performance of RFE for 2000 features can be explained by the fact that features are selected specifically for the classification task, while the autoencoder simply compresses the whole feature vector (which could notably exacerbate differences between sites), while at the same time being able to handle smaller database size contrarily to neural network models like the MLP.  
In the remainder of our experiments, we therefore use the RFE strategy with $C=2000$ features. 

% 250 features
% \begin{itemize}
% \item MLP: 66.5\%, AUC: 72\%, ensemble: 67\%
% \item AE: 66\%, AUC: 71\%, ensemble: 67.5\% 
% \item ridge: 65.5\%, AUC: 70\%, ensemble: 66\%  
% \item PCA: 53.2\%, AUC: 55.6\%, ensemble: 55.7\%
% \end{itemize}

% 500 features
% \begin{itemize}
% \item AE: 67.6\%, AUC: 73.4\%, ensemble: 68.8\%
% \item ridge: 67\%, AUC: 72.7\%, ensemble: 67.3\%
% \item MLP: 66.2\%, AUC: 72\%, ensemble: 66.7\%
% \item PCA: 50\%, AUC: 48\%, ensemble: 52.5\%
% \end{itemize}

% 1000 features
% \begin{itemize}
% \item AE: 68.2\%, AUC: 73.4\%, ensemble: 68.3\%
% \item ridge: 67.8\%, AUC: 73.6\%, ensemble: 68.2\%
% \item MLP: 65.5\%, AUC: 70.5\%, ensemble: 66.4\%
% \end{itemize}

% 2000 features
% \begin{itemize}
% \item ridge: 69.5\%, AUC: 75\%, ensemble: 70\%
% \item AE: 68.5\%, AUC: 73\%, ensemble: 68.8\%
% \item MLP: 65.8\%, AUC: 71.6\%, ensemble: 66\%
% \end{itemize}

\subsection{Influence of polynomial order K}
In a separate experiment we explored the influence of the Chebyshev polynomial order, $K$, which was empirically fixed to $K=3$ in our previous work for both ABIDE and ADNI databases.~\cite{kipf2016semi} tested the performance of ChebNet on three citation networks for $K \in \{1,2,3\}$ and found that a different degree led to optimal performance for each dataset. We believe that this difference in performance with regards to $K$ is related to intrinsic properties of the graphs. Here, we tested $K \in \{1,2,3,4,5\}$, which corresponds to filters learned for neighbours $K$-hops away from the node at the centre of the receptive field.

\begin{figure}[t]
\centering
\includegraphics[width=0.8\textwidth]{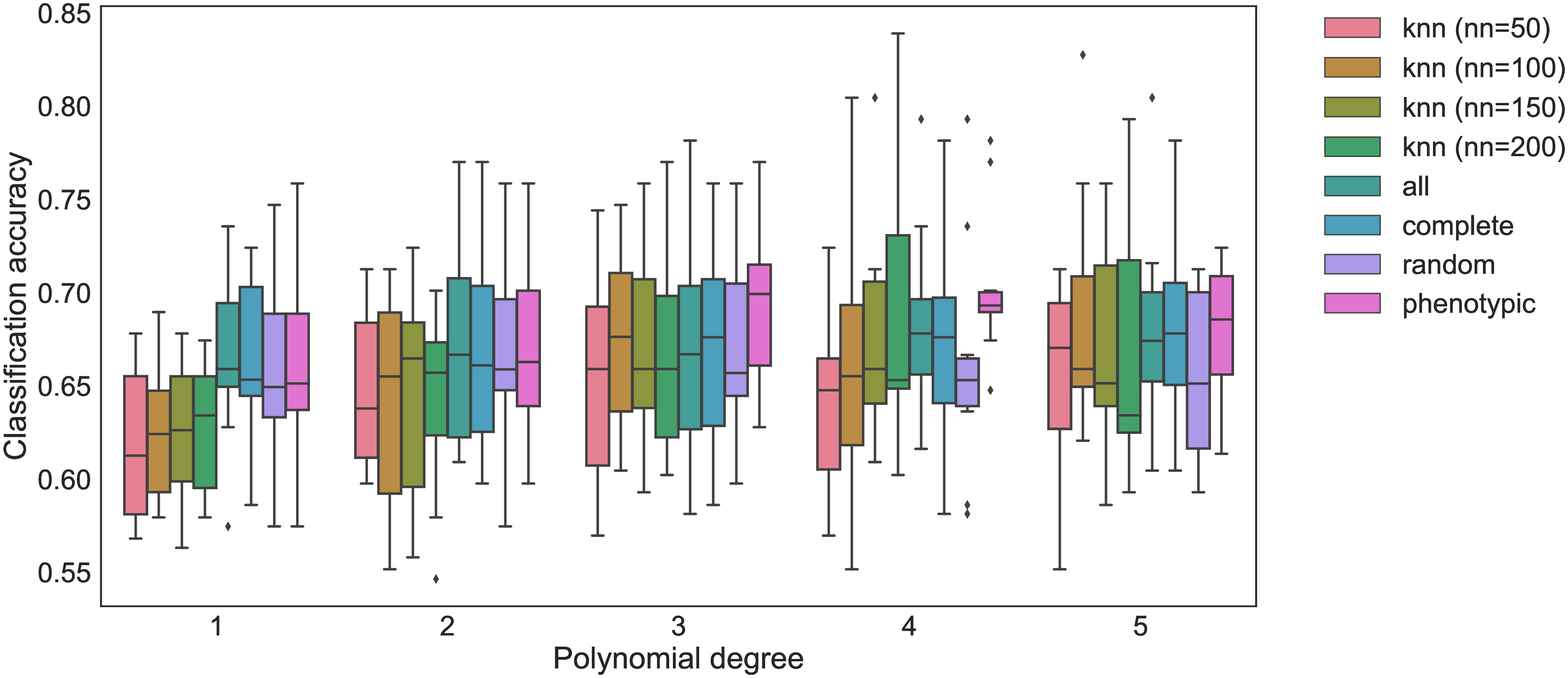} \\
\includegraphics[width=0.8\textwidth]{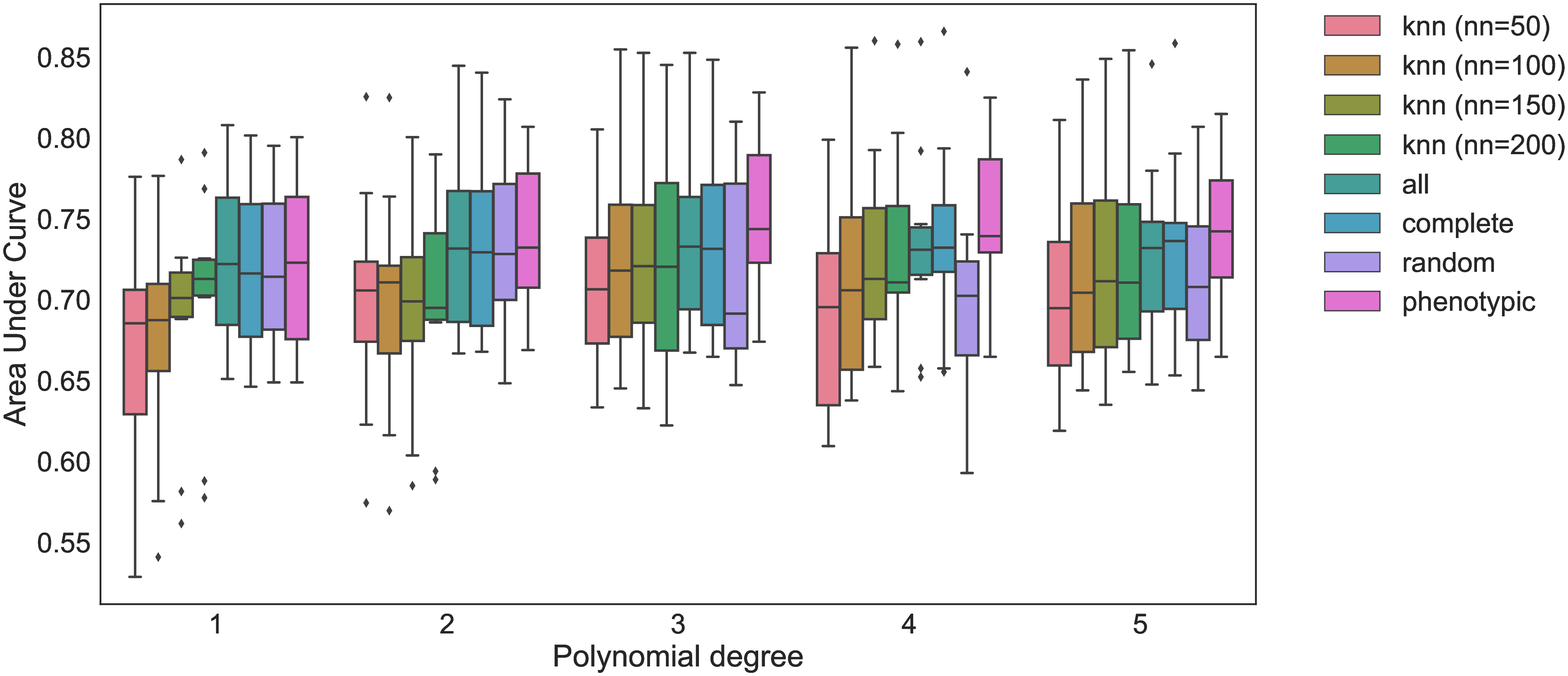}
\caption{Comparison of baseline graph structures and polynomial degrees for ABIDE dataset. Boxplots are generated for the performance (classification accuracy and area under curve) of each graph structure across 10 folds.}
\label{fig:abide_baselines}
\end{figure}

Boxplots for mean classification performance and area under curve across 10 folds on the ABIDE database are illustrated in Fig.~\ref{fig:abide_baselines}. We focus our attention to the \textit{phenotypic} graph (using default phenotypic measures), which corresponds to the graph constructed taking into account SEX and SITE information, weighted by the feature correlation similarity.  This graph structure was also chosen in~\cite{Parisot2017}. The best average performance is achieved for $K=4$ with accuracy 70.4\% and AUC 0.75. However, this is only marginally better than the average performance achieved with $K=3$, where accuracy reaches 69.5\% and same AUC. Performance seems to increase with $K$, starting with mean accuracy 65.8\% for $K=1$, but then drops to 67.9\% for $K = 5$.

\begin{figure}[t]
\centering
\includegraphics[width=0.8\textwidth]{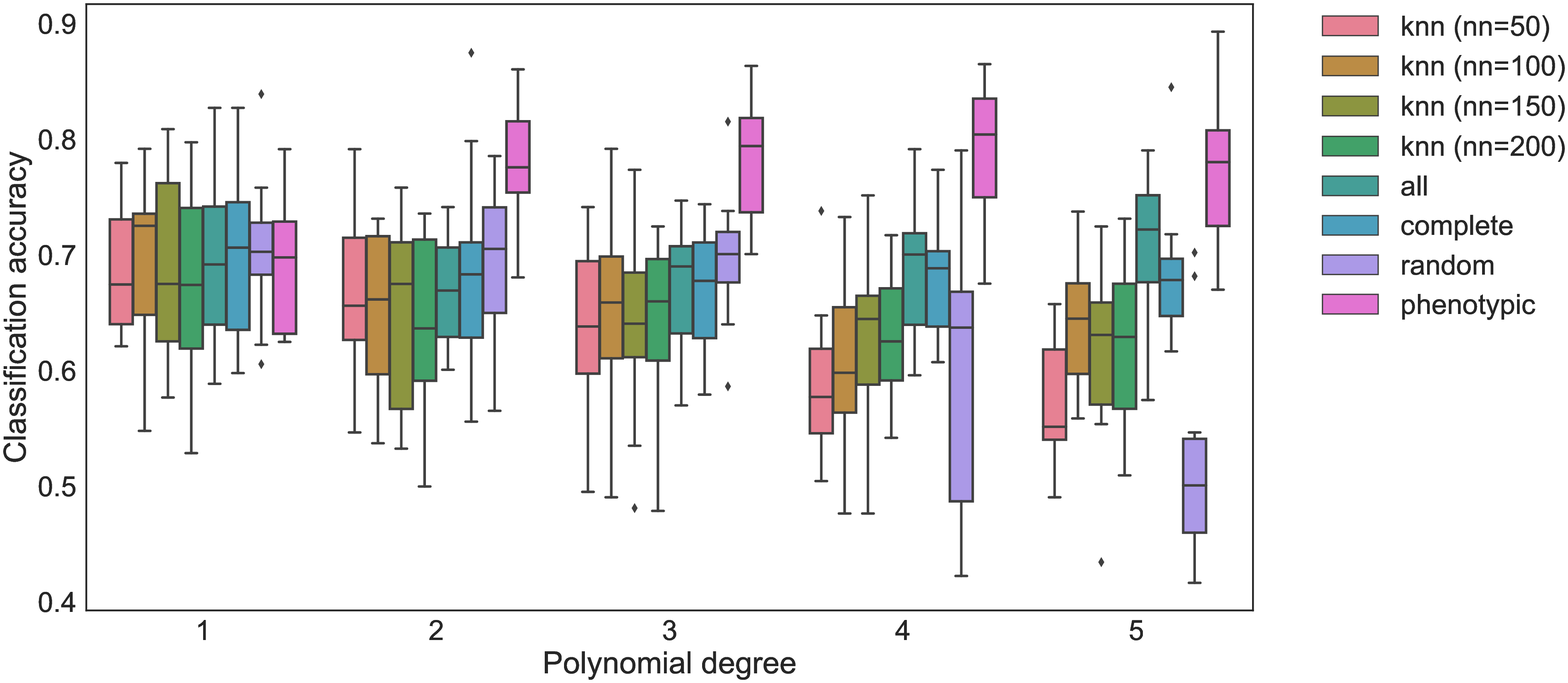} \\
\includegraphics[width=0.8\textwidth]{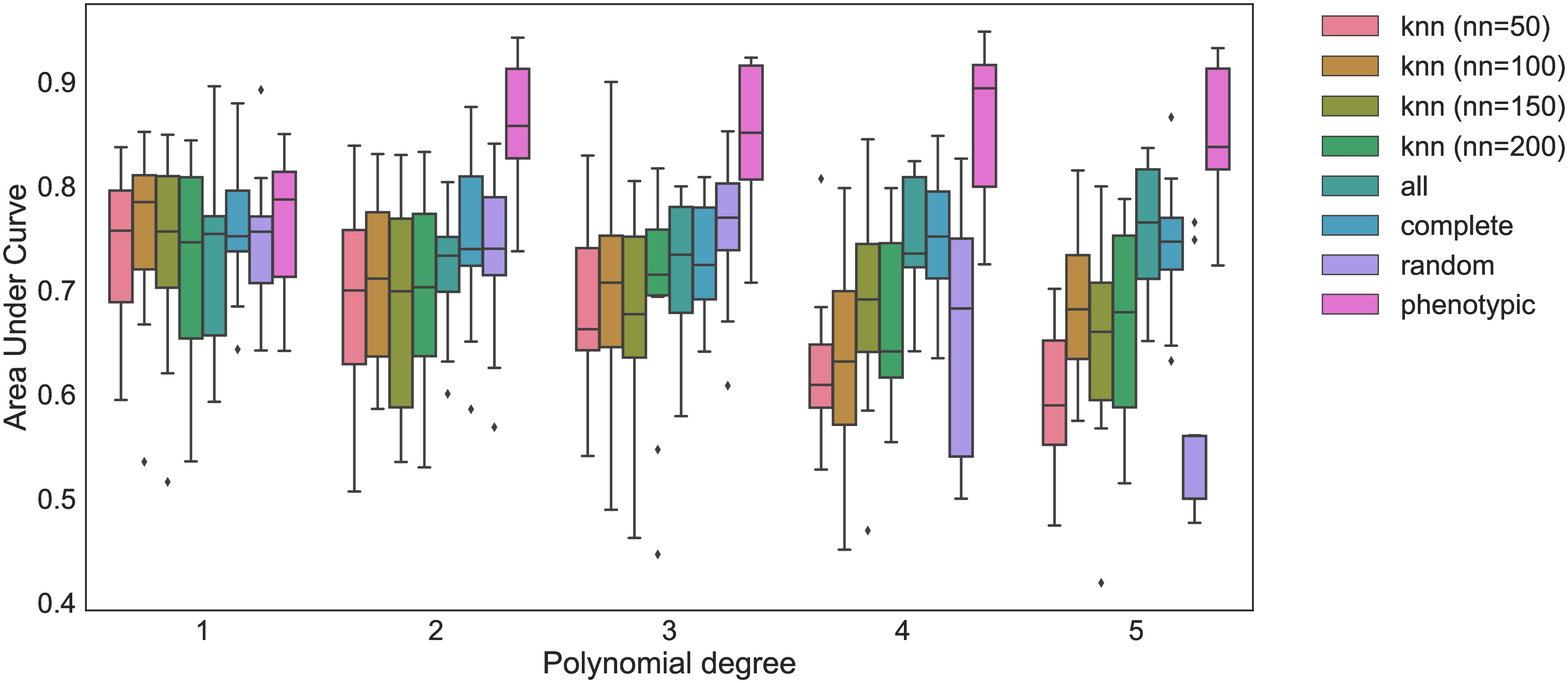}
\caption{Comparison of baseline graph structures and polynomial degrees for ADNI dataset. Boxplots are generated for the performance (classification accuracy and area under curve) of each graph structure across 10 folds.}
\label{fig:ADNI_baselines}
\end{figure}

A similar pattern is observed for the ADNI database, with results being summarised in Fig.~\ref{fig:ADNI_baselines}. Focusing again on the \textit{phenotypic} graph based on longitudinal information and previously used in~\cite{Parisot2017}, we can see that minimum performance is achieved with $K=1$ yielding average classification 69.3\% and AUC 0.76. Performance increases with $K$, leading to 78.3\% accuracy and 0.84 AUC for $K=3$, while for $K=4$ mean accuracy across 10 folds is 78.8\% and AUC is 0.86. Similarly to the behaviour mentioned above for the ABIDE database, performance drops to 77.7\% accuracy and 0.85 AUC for $K=5$ and decreases even further for higher values of $K$. All the above indicate that the receptive field is still limited for $K=1$, while for $K=3$ or $K=4$ it expands enough to capture the neighbourhood structure around a node. Higher values of $K$ likely enhance overfitting issues that arise from the limited amount of data used in comparison to tens of thousands of nodes deployed in other applications. The relationship between the optimal degree $K$ and the diameter of the graph is yet to be explored and can lead to insightful conclusions about the degree that leads to better performance for different graph structures.

\subsection{Graph construction strategy}
\label{sec:graph_cons}
As previously mentioned, since convolutions are parameterised on the Laplacian, the graph structure is expected to have a strong impact on semi-supervised classification performance. Therefore, we investigate the effect of different graph structures on average classification accuracy and area under curve across 10 folds. The explored graph structures include: \textbf{(a)} a k-nearest neighbours ($knn$) graph with $k=50, 100, 150, 200$, where (dis)similarities between nodes are based solely on their associated feature information, \textbf{(b)} a binary \textit{complete} graph, i.e. every node is connected to every other node with weight 1, \textbf{(c)} a complete graph weighted by the feature similarity between nodes (\textit{all}), \textbf{(d)} a \textit{random} graph with same edge density as the phenotypic graph but randomly rewired edges and \textbf{(e)} the default \textit{phenotypic} graph structure where non-imaging information is used along with the feature similarity to construct the graph.

For the ABIDE dataset, performance for the different graph structures is summarised in Fig.~\ref{fig:abide_baselines} along with the influence of the polynomial order for each for these structures. As we can observe, the $knn$ graphs are performing worse than \textit{all}, \textit{complete}, \textit{random} and \textit{phenotypic} for $K=1$ with average accuracy 61.8\% for $k=50$ and 62.8\% for $k=200$. Additionally $all$ and $complete$ graphs yield equivalent results in terms of classification accuracy (66.5\%), slightly outperforming the $random$ (65.6\%) and $phenotypic$ (65.8\%) graphs and indicating that for the first-order polynomial, edge density is more important than the quality of the graph. However, the \textit{phenotypic} graph is the one leading to the best performance compared to all other graphs for $K=3$ and $K=4$ (as wells as best overall) followed by the \textit{complete} graph (67.3\%) for $K=3$ and \textit{all} graph for $K=4$ (68.3\%). Interestingly, the performance achieved with the \textit{complete} graph is equivalent to that of a linear classifier, as presented later in~\ref{subsec:comp_other}, since all nodes contribute equally to filtering each node's features. The $phenotypic$ graph is also outperforming all alternative graph structures for every $K$ (except for $K=1$) by means of AUC, with the highest average AUC (0.75) achieved for $K=3$. It is also worth mentioning that the \textit{random} graph and the \textit{knn} graphs are the ones leading to the worst performance for $K \geq 2$, since they fail to capture the complex associations/similarities between all available subjects and, thus, do not help leverage the power of graph convolutions.

The equivalent results for the ADNI database are presented in Fig.~\ref{fig:ADNI_baselines}. These results are slightly different from the ones presented above for the ABIDE database, since the \textit{phenotypic} graph in this case is solely based on the longitudinal information and does not rely on the node features at all. Therefore, we observe a clearer `superiority' of the \textit{phenotypic} graph against all other graph structures for $K \geq 2$. Apart from that, there is a negative trend in terms of classification performance for the $knn$ and $random$ graphs with increasing $K$, demonstrating that irrelevant graphs or graphs that eliminate connections between subjects yield worse performance for high $K$ values. The best overall mean classification accuracy is achieved with the \textit{phenotypic} graph for $K=4$ (78.8\%) with AUC 0.86, followed by the \textit{all} (feature-based) graph with 69.0\% accuracy and 0.75 AUC. The \textit{complete} graph is also performing slightly worse than \textit{all} graph (accuracy 68.0\%, AUC 0.75), highlighting the fact that similarity-based edge weights are meaningful and help improve the quality of label propagation.

Across the two databases, we can observe that the meaningful \textit{phenotypic} graph leads to the best performance compared to all baseline graph structures. This pattern is more prevalent for the ADNI database, in which case the \textit{phenotypic} graph is feature-independent. The \textit{complete} and \textit{all} graph structures succeed the \textit{phenotypic} graph, especially for higher order polynomial filters, indicating that this population-based framework can benefit from interactions between all nodes, even if the structure is not optimal.

\subsection{Influence of the phenotypic measures}

We showed in Sec. \ref{sec:graph_cons} that phenotypic measures (the default measures successfully used in our previous work \citep{Parisot2017}) led to the best graph construction when compared with alternative structures. In this section, we evaluate the performance of the model for different phenotypic graph configurations. Considering 3 different measures and a similarity function for each database, we investigate the influence of each measure on the overall classification performance by training multiple GCNs with different combinations of such measures and similarities. Results are reported for multiple initialisation seeds, so as to investigate the stability of the method with respect to the graph structure. 

\begin{figure}[t]
\centering
\subfloat[Accuracy]{\includegraphics[width=0.7\textwidth]{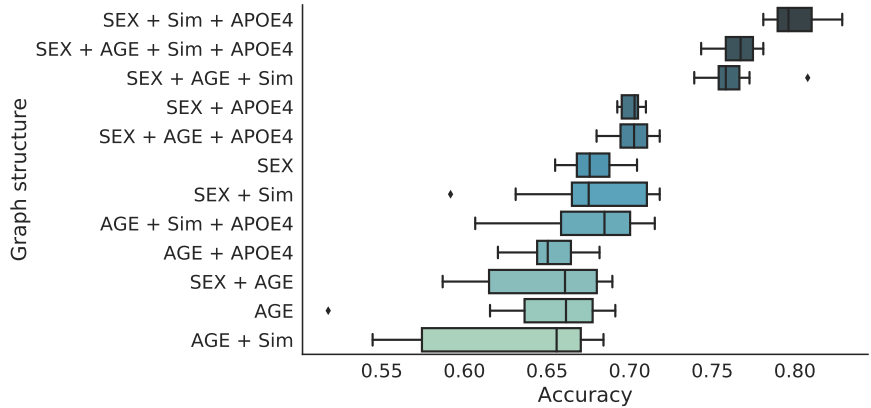}} \\
\subfloat[Area Under Curve]{\includegraphics[width=0.7\textwidth]{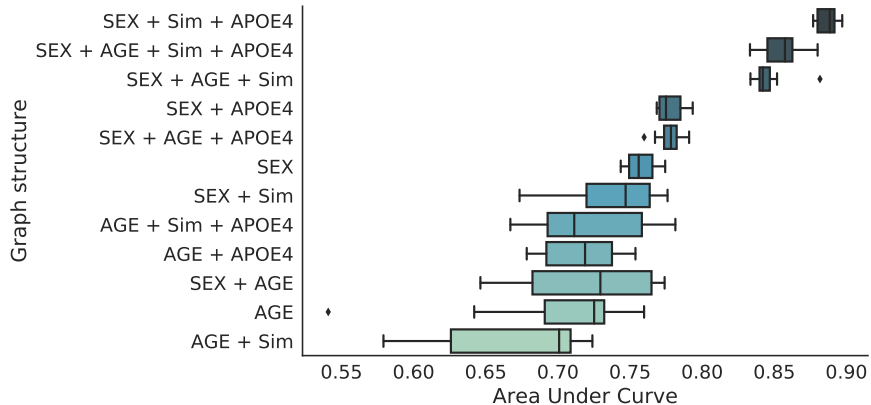}}
\caption{Influence of the phenotypic graph structure on the classification results on the ADNI database. The boxplots report the classification accuracy across 10 different initialisation seeds.}
\label{fig:ADNI_phen}
\end{figure}

The default graph structure for ADNI uses the similarity link between same subjects, SEX and AGE. We, furthermore, integrate genetic information regarding the APOE4 gene. Results for multiple graph configurations using Similarity, sex, age or APOE4 measures are reported in Fig. \ref{fig:ADNI_phen}.  
The first observation is a sharp decrease of stability across seeds (i.e. sensitivity to initialisation/lack of convergence) as the overall performance of the graph decreases. We observe a 14\% difference for worse performing graph and 5\% for best performing graph. This suggests that a poor graph structure significantly decreases robustness and convergence. This is to be expected as an inadequate graph structure leads to defining inaccurate neighbourhood systems, which can strongly impact the accuracy of the convolutions. 

Generally, we observe that performance worsens when using AGE as a phenotypic measure, while all other measures increase performance. The similarity between same subjects significantly increases accuracy (+10\% between SEX+APOE4 and SEX+APOE4+Sim), but worsens when using poor graph structures (e.g sex or age only). 
Last but not least, we can see that adding the APOE4 measure increases results, leading to 2 graph structures that beat the performance of the default graph. Our best performance is obtained for Similarity + SEX + APOE4, corresponding to a 80\% accuracy and 0.89 AUC. 

 \begin{figure}[t]
\centering
\subfloat[Accuracy]{\includegraphics[width=.7\textwidth]{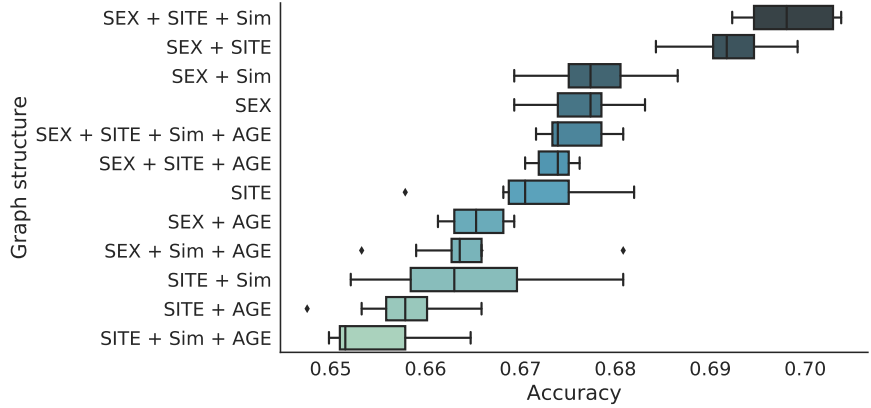}} \\
\subfloat[Area Under Curve]{\includegraphics[width=0.7\textwidth]{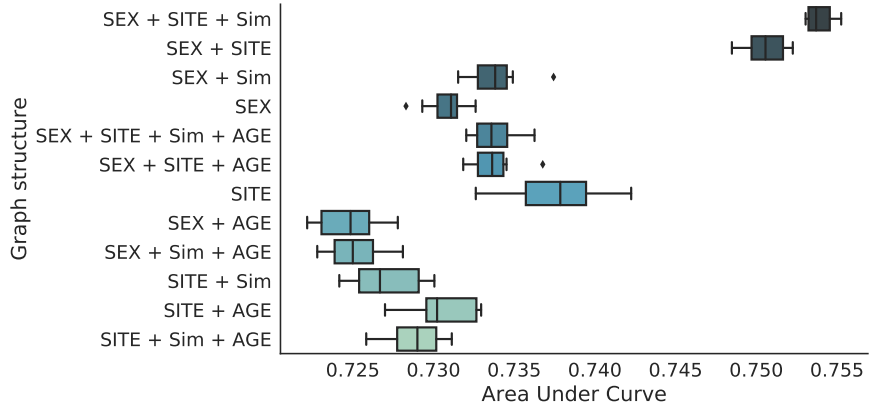}}
\caption{Influence of the phenotypic graph structure on the classification results on the ABIDE database. The boxplots report the classification accuracy across 10 different initialisation seeds.}
\label{fig:ABIDE_phen}
\end{figure}

Results for the ABIDE database are reported in Fig. \ref{fig:ABIDE_phen}. The default ABIDE graph is based on similarity between connectivity networks, SEX and SITE. In this experiment, we also introduce the AGE parameter.  
Contrarily to the ADNI database, we observe very little variation between graph structures with a $3\%$ difference in accuracy between best and worse performing graphs. Similarly, the difference between different initialisation seeds is almost negligible.  
The best performing graph is the one used in \cite{Parisot2017}, i.e. the default graph, with an average accuracy over 10 seeds of 0.70 and 0.75. The similarity provides a small increase for almost all cases. SEX appears to provide a better accuracy than SITE, possibly due to the fact that the SITE based graph is highly disconnected. However, SITE provides better results in terms of AUC.  
Similarly to what was observed with ADNI, AGE consistently reduces the classification performance. This observation on both databases may be linked to the way that it is integrated in the graph, as it is the only phenotypic quantitative measure used in this paper. 

\begin{figure}[t]
\centering
\subfloat[ABIDE]{\includegraphics[width=0.75\linewidth]{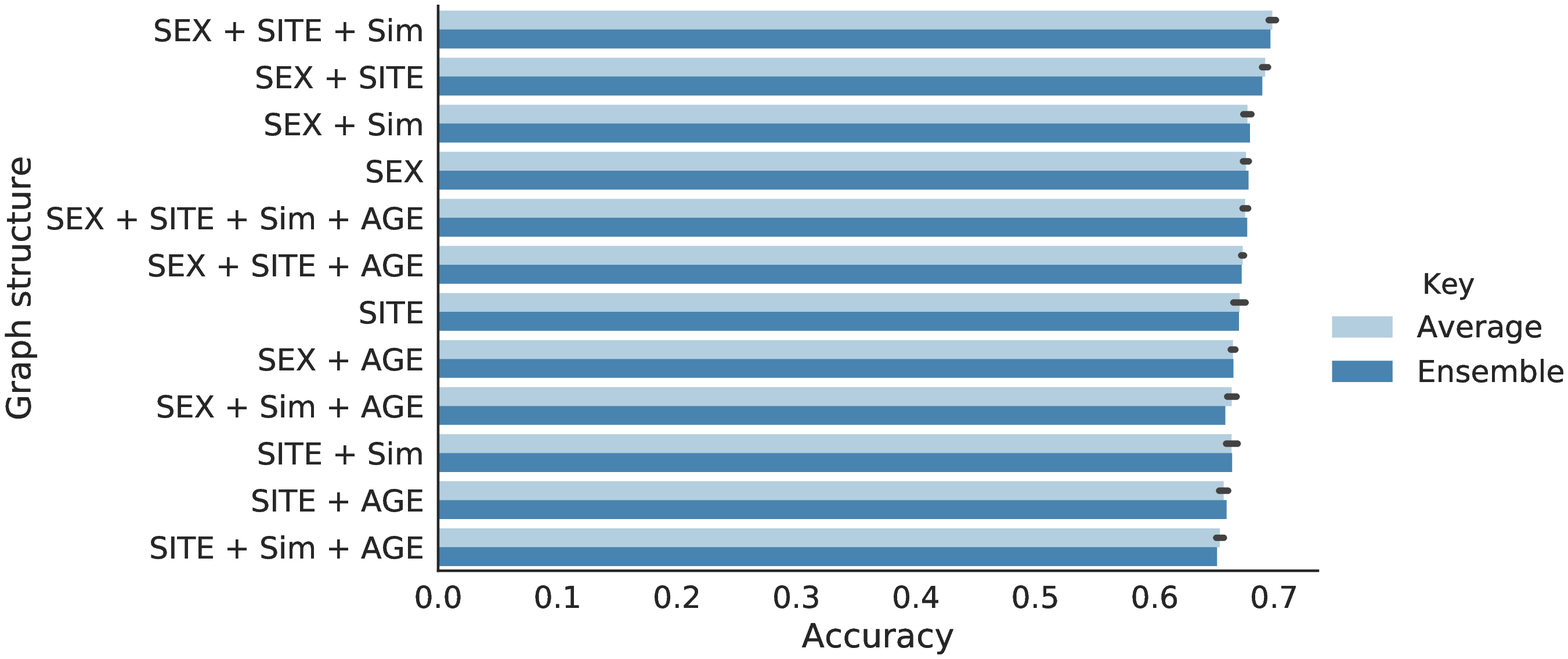}} \\
\subfloat[ADNI]{\includegraphics[width=0.75\linewidth]{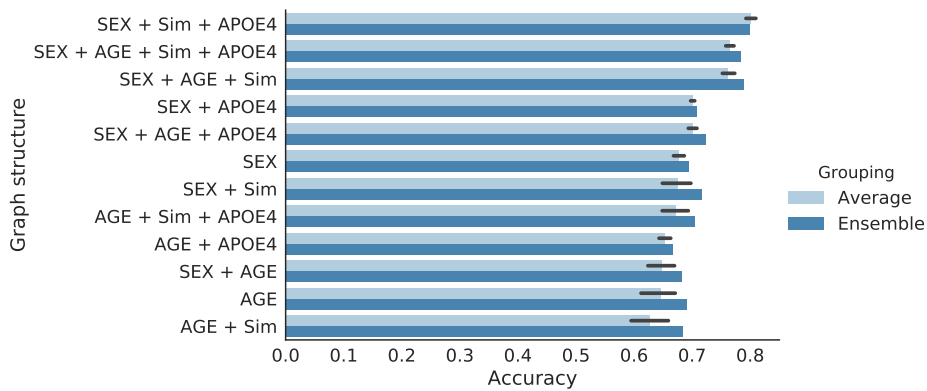}}
\caption{Classification accuracy for the ABIDE and ADNI database. Results are reported across 10 different initialisation seeds, either by averaging results across seeds (light blue), or by ensembling all results (dark blue).}
\label{fig:ensemble}
\end{figure}

Finally, we compare two strategies to evaluate the overall performance between seeds: 1) averaging the accuracies obtained for all seeds and 2) ensembling the results using majority voting. We provide comparative results in Fig. \ref{fig:ensemble} for both ADNI and ABIDE database. We can see that there is very little difference between both strategies for the ABIDE database and the best performing graphs of the ADNI database, while a significant increase in performance is obtained using ensembling of poor quality graphs which tend to have very unstable performances across seeds. The best performance is almost consistently obtained by ensembling. This suggests that the optimal strategy is to ensemble between multiple initialisations. However, the small variability between initialisations for good graph structures suggests that results are consistent across seeds and that one initialisation provides sufficient evaluation.

\subsection{Comparison to other methods}
\label{subsec:comp_other}

% \sarah{could be useful to add a table with ADNI reported results from other methods? different data, but could be added to show that we are aware of the literature...could ask Ricardo?}

\begin{figure}[t]
\centering
\subfloat[ABIDE accuracy]{\includegraphics[width=0.4\textwidth]{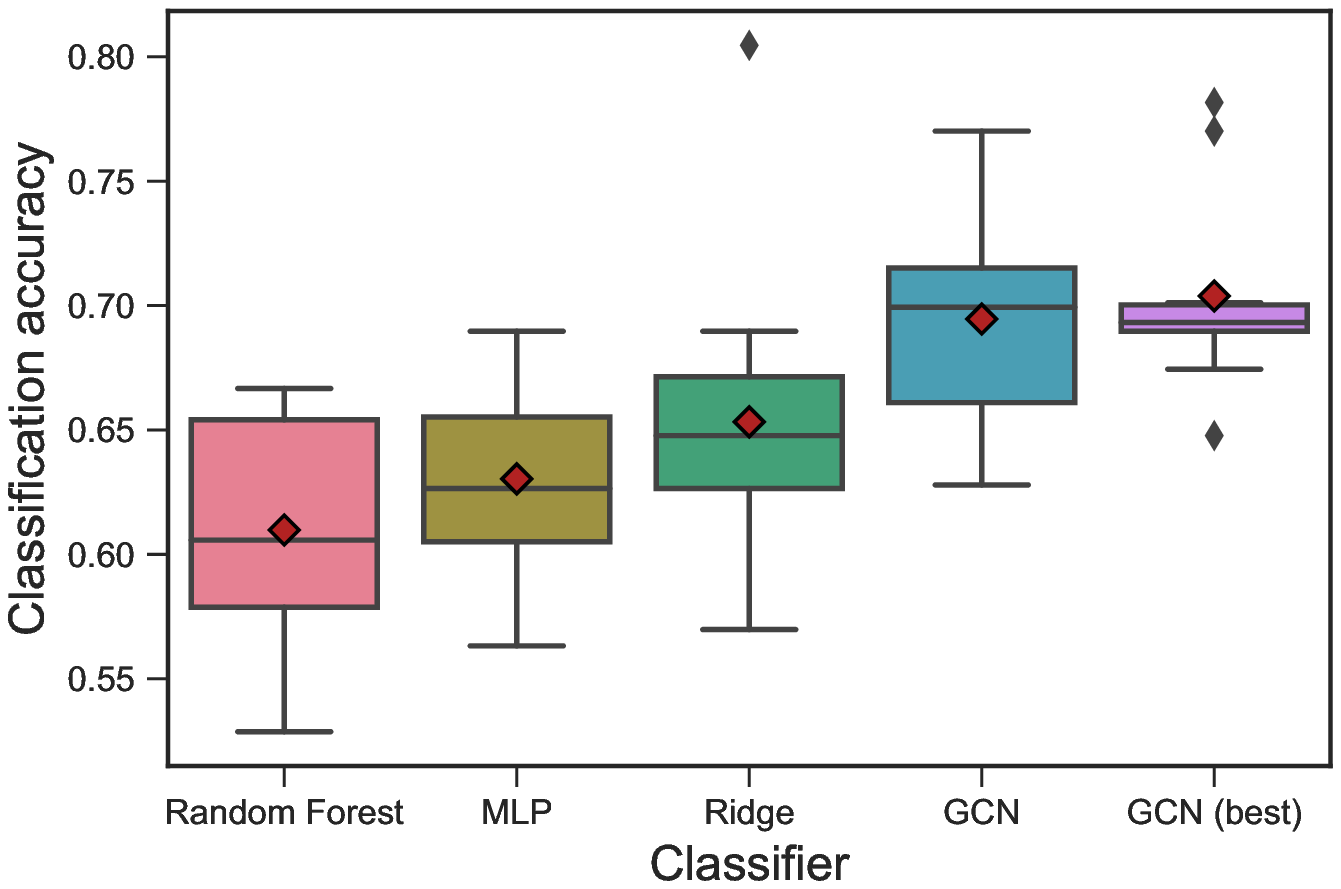}\label{subfig:ABIDEacc}}
\subfloat[ABIDE AUC]{\includegraphics[width=0.4\textwidth]{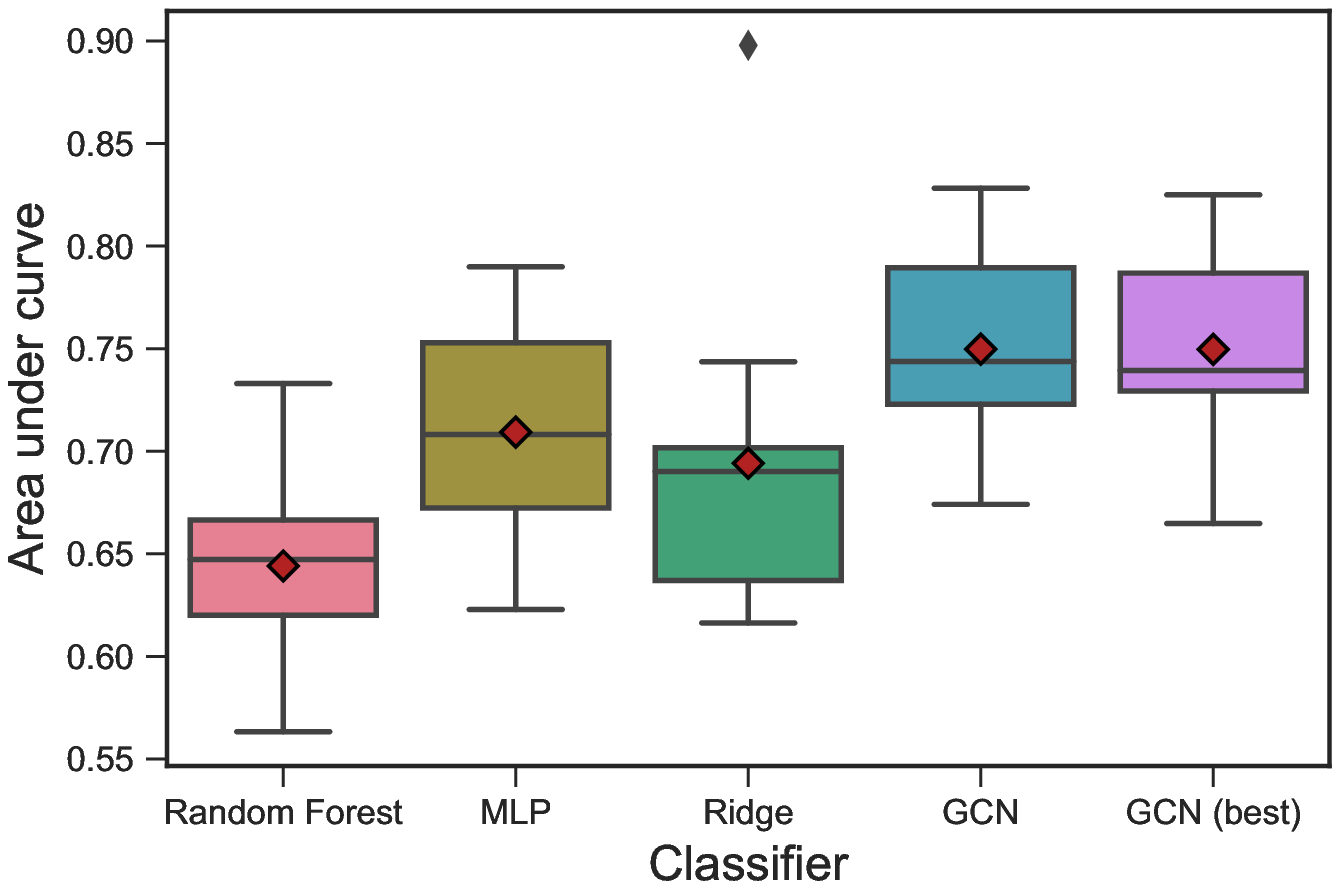}\label{subfig:ABIDEauc}} \\
\subfloat[ADNI accuracy]{\includegraphics[width=0.4\textwidth] {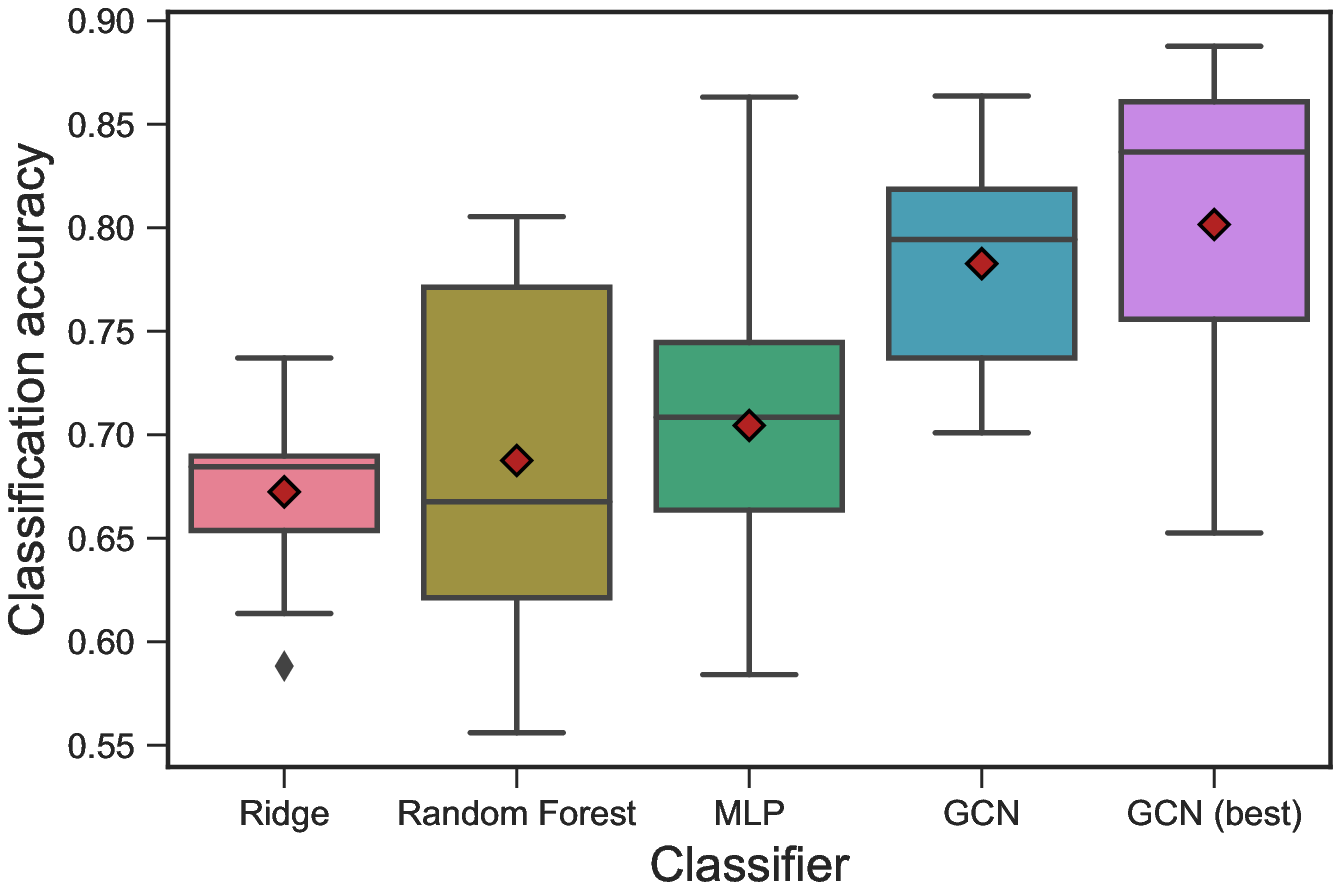}\label{subfig:ADNIacc}}
\subfloat[ADNI AUC]{\includegraphics[width=0.4\textwidth]{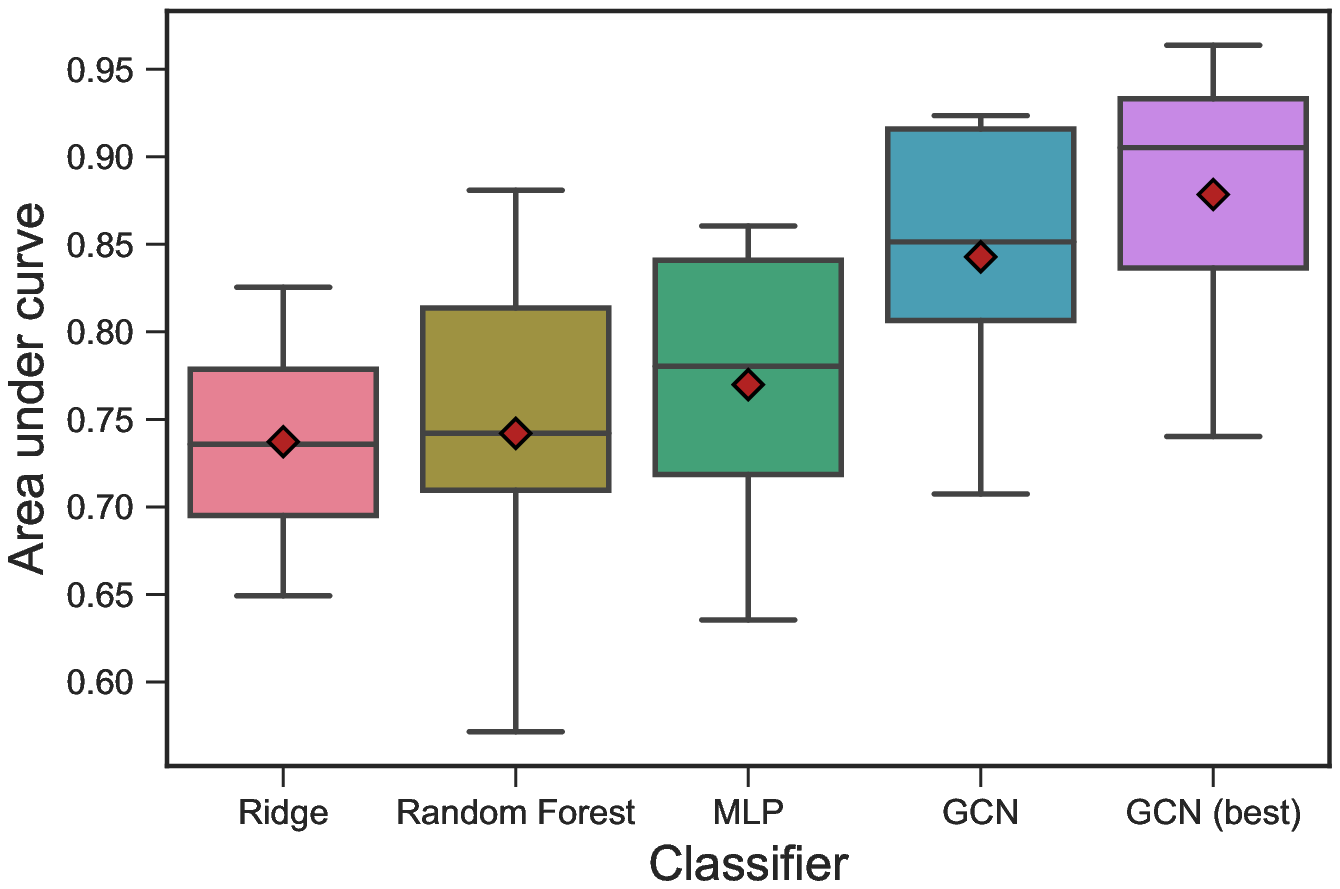}\label{subfig:ADNIauc}}
\caption{Comparative boxplots of the classification accuracy and area under curve (AUC) over all cross validation folds for the (a, b) ABIDE and (c, d) ADNI databases (MCI conversion task) for different baseline classifiers (random forest, ridge and MLP), the GCN model used in~\cite{Parisot2017} and the best GCN model from the current extensive evaluation.}
\label{fig:boxplots}
\end{figure}

Apart from investigating how the different components of the proposed framework impact semi-supervised classification performance, we further compared the GCN results to different well-established classifiers. These include a ridge classifier (using the scikit-learn library implementation \cite{sklearn}), which showed the best performance amongst linear classifiers, a random forest classifier with 100 estimators and a multi-layer perceptron (MLP) classifier. We also add the performance reported in \cite{Parisot2017} using the GCN model (GCN) alongside the best performance obtained in this manuscript throughout our experiments (GCN (best)). This is to highlight the performance increase obtained in this paper, and for consistency purposes with \cite{Parisot2017}. For the MLP we used the same parameters as the GCN implementation, in terms of number of hidden layers, number of features, dropout, seed, learning rate and regularisation, and fixed the number of epochs to 200 for both datasets. Comparative boxplots across all folds between these three approaches, along with the previously used model and the best achieved result (in terms of accuracy) for the same seed through our extensive evaluation in this paper are shown in Fig. \ref{fig:boxplots} for both databases. 

The worst performance on the ABIDE database is observed for the random forest classifier with 61.0\% average accuracy and 0.64 AUC. The ridge classifier is doing better than the MLP in terms of classification accuracy (65.3\% vs 63.0\%), but is outperformed by the MLP in terms of AUC (0.69 vs 0.71). The best overall performance observed with the GCN model is 70.4\% accuracy and 0.75 AUC, outperforming the recent state of the art (66.8\%) \cite{abraham2016deriving}. For the ADNI database, the worst performance is achieved with the ridge classifier (67.2\% accuracy and 0.74 AUC), followed by the random forest classifier (68.8\% and 0.74 AUC). The MLP classifier yields slightly improved results compared to these two classifiers with 70.4\% mean classification accuracy and 0.77 AUC. GCN results for this database show a large increase in performance with respect to the competing methods, with an average accuracy of 80.0\% and AUC of 0.88, higher than state of the art results \cite{tong2016}, corresponding to a 9\% increase over an MLP. Finally, regarding the GCN model, we observe a 4\% increase in accuracy for the ADNI database using a better graph structure (integrating APOE4 gene information and eliminating AGE information) with respect to \cite{Parisot2017}. The ABIDE performance is stable, suggesting an optimal graph structure in \cite{Parisot2017} or a limitation inherent to the dataset. 

% For both databases, we observe a significant ($p<0.05$) increase both in terms of accuracy and area under curve using our proposed method, with respect to the competing methods. The random support yields equivalent or worse results to the linear classifier. For ABIDE, we obtain an average accuracy of 69.5\%, outperforming the recent state of the art (66.8\%) \cite{abraham2016deriving}. Results obtained for the ADNI database show a large increase in performance with respect to the competing methods, with an average accuracy of 77\% on par with state of the art results \cite{tong2016}, corresponding to a 10\% increase over a standard linear classifier. 

\section{Discussion}

% \sarah{TODO: substantially expand this, do we leave the fact that it's a proof of concept? discussion on limitations,  high variability between seeds, especially for suboptimal graph structures: more reliable/better performance obtained using ensembles in those cases; feature visualisation/interpretation; ability to handle smaller databases than regular neural network methods  }

In this paper, we proposed a method for group-level population diagnosis that exploits the novel concept of spectral graph convolutions. We modelled populations as a sparse graph combining subject-specific imaging data and pairwise interactions described using phenotypic and other non-imaging information. This sparse graph is used to train a GCN in a semi-supervised manner for node classification, learning on a subset of labelled nodes and evaluating on the rest. Our experiments on two large and challenging databases (ABIDE and ADNI) confirm our initial hypothesis about the importance of contextual pairwise information for classification,  as we obtain state of the art performance with 70.4\% (ABIDE) and 80\% (ADNI) accuracy, corresponding to increases of 5\% and 9\% with respect to classifiers using node features only. Our extensive evaluation analyses the different components of the model, including the feature selection method, polynomial degree and graph construction strategy. Exploring different graph structures and baselines, we show how our phenotypic graph formulation yields more accurate and stable results, as well as the importance of choosing appropriate phenotypic measures to model the pairwise interactions.

While our method is tested against multiple baselines, conditional random fields (CRF)/ Markov random fields (MRF) models also show interesting similarities with our approach. As in our setting, CRF models are cast as graph labelling problems and seek to increase classification performance by modelling pairwise interactions between graph nodes. Nonetheless, such approaches require substantial methodological decisions (definitions of node likelihood from a separate learning strategy, devising an optimisation strategy adapted to the considered graph structure and optimising parameters), making direct comparison not straightforward. Furthermore, CRF formulations model feature vectors and pairwise interactions sequentially, leading to a weaker model than our approach which learn end-to-end features representation informed by pairwise interactions. 

In this paper, we cast the problem of AD and ASD diagnosis as a binary classification due to the annotations provided in the complete ABIDE and ADNI databases. However, it is well known that both diseases lie on a spectrum \citep{mega1996spectrum,volkmar2009autism}. One could therefore be seeking to carry out multi-class classification or to predict  continuous outputs in such setting. This can be done easily using our framework by updating the output size and loss function to predict multi-class labels or carry out a regression task.

% \sarah{TODO rewrite paper summary}
% In this paper, we introduced the novel concept of graph convolutions for population-based brain analysis. We proposed a strategy to construct a population graph combining image based patient-specific information with non-imaging based pairwise interactions, and use this structure to train a GCN for semi-supervised classification of populations. As a proof of concept, the method was tested on the challenging ABIDE and ADNI databases, respectively for ASD classification from a heterogeneous database and predicting MCI conversion from longitudinal information. Our experiments confirmed our initial hypothesis about the importance of contextual pairwise information for the classification process. In the proposed semi-supervised learning setting, conditioning the GCN on the adjacency matrix allows to learn representations even for the unlabelled nodes, thanks to the supervised loss gradient information that is distributed across the network. This has a clear impact on the quality of the predictions, leading to about 4.1\% improvement for ABIDE and 10\% for ADNI when comparing to a standard linear classifier (where only individual features are considered). 

The ABIDE database is particularly challenging due to the fact that images are acquired at different sites with different protocols. This strongly reduces the comparability of image features from different sites. Our experiments suggest that  an accuracy of approx. $70\%$ could be an intrinsic limitation of the whole dataset. This observation aligns with the most recent works on the ABIDE dataset, reporting $68 \%$ \cite{abraham2016deriving} and $70\%$ \cite{heinsfeld2018} accuracies on the whole dataset (using leave one out cross validation for the latter). It is likely that increasing performance on this dataset would require models that can learn to eliminate this inter-site variability by capturing site-invariant class-discriminative patterns. It should be noted that significantly better accuracy results can be obtained when restricting analysis on one site only or when integrating cognitive test results.

Spectral GCNs provide a powerful and principled way of performing convolutions on irregular graph structures. In this paper, we use Chebyshev polynomials, which have proven to be an excellent compromise between speed and accuracy. The recently introduced Cayley polynomials \citep{levie2017cayleynets} have shown a lot of potential and could provide further improvements of our results. Despite their many advantages, spectral GCNs have an important drawback, which is that they can only be applied to graphs of fixed structure due to their parametrisation on the graph Laplacian. While a small modification of the graph structure (e.g. replacing a node) is unlikely to alter performance, retraining the GCN will be a necessity for substantially modified graphs. This also poses a challenge when new subjects become available and need to be incorporated in the analysis, as the model will need to be trained from scratch. In situations where one expects highly variable graph structures, spatial GCNs \citep{monti2016geometric} should be considered.

Another limitation is the use of hand crafted features as node descriptors. The main advantage of deep learning is that the network learns itself an optimal feature representation of the raw input data for the task at hand. Ideally, one would want an end to end strategy, learning optimal node features from the input images or connectivity networks. This could potentially be achieved using the method proposed in \cite{monti2017completion} for matrix completion, proposing a multi-graph convolution method. In our case, the second graph could be the input image data or connectivity networks.
Despite their limited performance in our experiments, feature encoding using autoencoders or MLPs have the potential of providing better results than RFE. Here we use very simple models for both AE and MLP; therefore, their performance is likely to increase with appropriate parameter and model engineering. Nonetheless, the main limitation of those neural network based approaches compared to RFE is the limited amount of training data, especially for the smallest sites. Using additional data (for example using the recently released ABIDE II dataset) has the potential to increase the performance of such methods, and could also allow using adversarial techniques to learn a feature representation that is independent of the acquisition site. Last but not least, the weak performance of the considered learning based methods (AE, MLP, PCA) could be due to an excessing amount of data learning. Learning a representation from existing relationships within the dataset, and repeating the same process in the GCN network, amounts to double learning the data which could reduce performance. On the contrary, RFE does not learn a new representation of the feature vector.

Devising an effective strategy to construct the population graph is essential and far from obvious. We have explored several graph structures in this paper and demonstrated how the graph can significantly affect classification accuracy. Our phenotypic graph construction strategy yielded the best performance, nonetheless, each graph edge comprises multiple types of information. Our experiments have shown that integrating the wrong or redundant phenotypic information (e.g. patient age in our case) has a strong negative impact on the results, while accurate measures with known links to the pathologies substantially increase performance. This is in accordance with our initial hypothesis that integrating non-imaging data that is known to influence the imaging data or subject's label yields better feature representations. 

Another graph construction strategy could be a pure learning based approach, learning a graph structure from self-attention weights and using all potential phenotypic measures as features. An added value of such an approach could be the identification  of new important phenotypic measures by exploration of learned attention weights. We could also consider attributed graph edges (i.e. a vector instead of a scalar weight) comprising all measures. Such a model could be exploited using a recent spatial GCN strategy \citep{simonovsky2017dynamic}. The proposed framework could benefit from more sophisticated graph representation learning techniques, which can help the discovery of edge features and contribute to improved transfer-learning and node classification results~\citep{rossi2017deep}. One could also build multiple graph structures (e.g. one per phenotypic measure, or a mixture of phenotypic and knn graphs) and find a common Fourier base for convolutions via joint diagonalisation of the graphs' Laplacian matrices \citep{eynard2015multimodal}.

% Several extensions could be considered for this work. Devising an effective strategy to construct the population graph is essential and far from obvious. Our graph encompasses several types of information in the same edge. An interesting extension would be to use attributed graphs, where the edge between two nodes corresponds to a vector rather than a scalar. This would allow to exploit complementary information and weight the influence of some measures differently.

%Integrating time information with respect to the longitudinal data could also be considered. 
%Our feature vectors are currently quite simple, as our main objective was to show the influence of the contextual information in the graph. We plan to evaluate our method using richer feature vectors, potentially via the use of autoencoders from MRI images and rs-fMRI connectivity networks. 
Among the main limitations of this work one should consider the generalisation of this framework to unseen sites, e.g. in the ABIDE case. Since this is an application of transductive learning, generalisation to new unseen domains is expected to lead to performance decrease, especially if the training dataset is not large enough to capture population variability. Moreover, the way the ADNI graph is constructed, with multiple scans per subject being modelled as nodes and classified independently, is likely to introduce biases towards subjects with more visits available.
Last but not least, highly class imbalanced problems constitute a scenario that would require further research. In this work, we performed two studies with relatively balanced data. However, in certain types of population studies (e.g. genome-wide predictions tasks \cite{Jones2017}) one can find huge
class imbalance ratios, in the order of 1:10000. In the future, we would like to study how graph convolutions can be used to leverage the available annotated data towards improving prediction rate in highly class imbalanced problems.

\section*{Acknowledgements}
This work was supported by the European Union's Seventh Framework Programme (FP/2007-2013) / ERC Grant Agreement no. 319456. The Titan X Pascal used for this research was donated by the NVIDIA Corporation. Enzo Ferrante is beneficiary of an AXA Research grant. Sofia Ira Ktena is supported by the EPSRC Centre for Doctoral Training in High Performance Embedded and Distributed Systems (HiPEDS, Grant Reference EP/L016796/1).

\section*{References}

\bibliographystyle{model2-names}
\bibliography{references}

\end{document}